\pdfoutput=1

\documentclass[11pt]{article}

\usepackage[final]{acl}
\usepackage{amsmath}
\usepackage{hyperref}
\usepackage{listings}
\usepackage{fancyvrb}
\usepackage{enumitem}

\usepackage{times}
\usepackage{latexsym}
\usepackage[breakable]{tcolorbox}

\usepackage[T1]{fontenc}

\usepackage[utf8]{inputenc}

\usepackage{microtype}

\usepackage{inconsolata}
\usepackage{array}
\usepackage{xcolor}

\usepackage{graphicx}
\usepackage{subcaption}

\usepackage{tabularx}
\usepackage{booktabs}

\usepackage{tcolorbox}
\usepackage{listings}
\tcbuselibrary{skins,breakable}

\tcbset{
    promptbox/.style={
        colback=gray!5!white,
        colframe=gray!40!black,
        arc=2mm,
        boxrule=0.5pt,
        left=5mm,
        right=3mm,
        top=3mm,
        bottom=3mm,
        fonttitle=\bfseries,
        breakable
    }
}

\title{The Pragmatic Mind of Machines: Tracing the Emergence of Pragmatic Competence in Large Language Models}

\author{
Kefan Yu\textsuperscript{$\dagger$}\thanks{Both authors contributed equally. Correspondence to \texttt{qcz@u.northwestern.edu}},
Qingcheng Zeng\textsuperscript{$\dagger$}\footnotemark[\value{footnote}],
Weihao Xuan\textsuperscript{$\ddag,\diamond$},
Wanxin Li\textsuperscript{$\sharp$},
\textbf{Jingyi Wu}\textsuperscript{$\dagger$},
\textbf{Rob Voigt}\textsuperscript{$\mathsection$}\\\\
\textsuperscript{$\dagger$}Northwestern University \quad
\textsuperscript{$\ddag$}The University of Tokyo \quad
\textsuperscript{$\diamond$}RIKEN AIP \\
\textsuperscript{$\sharp$}Zhejiang University \quad
\textsuperscript{$\mathsection$}University of California, Davis\\\\
\texttt{qcz@u.northwestern.edu}
}

\begin{document}
\maketitle
\begin{abstract}
Current large language models (LLMs) have demonstrated emerging capabilities in social intelligence tasks, including implicature resolution and theory-of-mind reasoning, both of which require substantial pragmatic understanding. However, how LLMs acquire this pragmatic competence throughout the training process remains poorly understood. In this work, we introduce \textsc{AltPrag}, a dataset grounded in the pragmatic concept of \textit{alternatives}, to evaluate whether LLMs at different training stages can accurately infer nuanced speaker intentions. Each instance pairs two equally plausible yet pragmatically divergent continuations and requires the model to (i) infer the speaker’s intended meaning and (ii) explain when and why a speaker would choose one utterance over its alternative, thus directly probing pragmatic competence through contrastive reasoning. We systematically evaluate 22 LLMs across 3 key training stages: after pre-training, supervised fine-tuning (SFT), and preference optimization, to examine the development of pragmatic competence. Our results show that even base models exhibit notable sensitivity to pragmatic cues, which improves consistently with increases in model and data scale. Additionally, SFT and RLHF contribute further gains, particularly in cognitive-pragmatic scenarios. These findings highlight pragmatic competence as an emergent and compositional property of LLM training and offer new insights for aligning models with human communicative norms.
\end{abstract}

\section{Introduction}
Human communication typically extends beyond the literal interpretation of utterances. Pragmatics, the branch of linguistics concerned with how context shapes meaning, is central to natural language understanding. It encompasses a range of phenomena such as implicature \cite{sadock1978testing}, presupposition \cite{karttunen1974presupposition}, and indirect speech acts \cite{searle1975indirect}.
\begin{figure}[h]
    \centering
    \includegraphics[width=0.925\columnwidth]{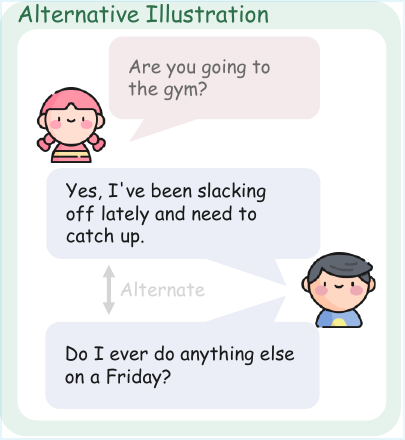}
    \caption{Illustration of alternatives. Two appropriate replies to the same question convey different pragmatic forces, the upper direct and explanatory, the lower playful and implicitly affirmative. We prompt LLMs to interpret the speaker’s intent behind each reply and articulate situational motivations one would be preferred over the other, thereby isolating pragmatic reasoning by holding the context and literal content constant.}
    \label{fig:alternative}
\end{figure}
With the advent of LLMs, a growing body of research has begun to explore whether these models exhibit sensitivity to pragmatic cues. Recent studies have investigated LLMs’ abilities to infer speaker intentions \cite{hu_fine-grained_2023, ruis, sravanthi-etal-2024-pub}, perform theory-of-mind reasoning \cite{doi:10.1073/pnas.2405460121, chen-etal-2024-tombench, shapira-etal-2024-clever}, and even pass Turing tests in controlled settings \cite{jones2025largelanguagemodelspass}. These findings hint at emergent pragmatic abilities in LLMs, motivating deeper inquiry.

However, it remains an open question at which stage of training LLMs acquire sufficient pragmatic understanding. \citet{ruis} conducted an empirical study showing that only example-level instruction-tuned (IT) models will significantly outperform random baselines on pragmatic tasks. Nonetheless, their evaluation faces two key limitations. First, their analysis is based on a binary classification task \cite{GEORGE20202316}, in which models only respond "yes" or "no" to specific utterances - an approach that may oversimplify the complexity of context and pragmatic reasoning. Second, the category of example-level IT models they examine primarily includes proprietary models such as GPT-3.5 and GPT-4, for which the specific training procedures are not publicly known. In particular, it is unclear when techniques such as reinforcement learning from human feedback (RLHF) are applied, thus making it difficult to draw firm conclusions about how pragmatic competence correlates with specific phases of training.

In this paper, we introduce \textsc{AltPrag}, a human-in-the-loop annotated dataset grounded in the notion of alternatives in pragmatics. As illustrated in \autoref{fig:alternative}, each dialogue instance pairs two equally valid but pragmatically distinct continuations, surfacing fine-grained differences in speaker intent and communicative strategy. Using this dataset, we ask models to infer the speaker intent behind each alternative to probe the pragmatic capabilities of LLMs at different training stages, specifically, after pre-training, SFT, and preference optimization. To evaluate model performance, we adopt an LLM-as-a-judge framework, comparing model-generated interpretations with human-verified references. Our results and contributions can be summarized as follows:
\begin{itemize} [itemsep=0.35pt]
    \item We present the first systematic analysis\footnote{
Code is available at \href{https://github.com/Huangtubaye233/PragmaticsLLM}{GitHub}.
Dataset is available at \href{https://huggingface.co/datasets/Huangtubaye233/AltPrag}{HuggingFace}.
} of how pragmatic competence evolves across different training stages of LLMs, using a free-form evaluation framework to capture fine-grained pragmatic judgments.
    \item We find that even base LLMs already exhibit measurable pragmatic competence, which scales with model size and training data volume—\emph{a result that contrasts with findings reported by} \citet{ruis}.
    \item We further show that both SFT and DPO help improve pragmatic understanding, especially capturing the cognitive-pragmatic nuances.
\end{itemize}

\section{Related Work}
\textbf{Pragmatics in LLMs.} The extent to which large
language models (LLMs) understand and process pragmatic phenomena has been the focus of increasing scholarly attention. A recent survey by \citet{DBLP:journals/corr/abs-2502-12378} reviews the rapid progress on LLM pragmatic abilities, cataloguing datasets, evaluation protocols, and open challenges. \citet{hu_fine-grained_2023} evaluated a range of LLMs and showed that the largest ones nearly match humans on deception, indirectness, and irony. Building on this line of inquiry, \citet{sravanthi-etal-2024-pub} released a benchmark covering subtler pragmatic reasoning beyond multiple-choice tests. Extending this evaluation paradigm further, \citet{wu-etal-2024-rethinking} proposed free-form pragmatic tasks and demonstrated that preference optimization may serve as a “free lunch” for enhancing pragmatic competence. Complementary work targets specific pragmatic phenomena with tailored probes. Reference-game setups test speaker–listener coordination \cite{shaikh-etal-2023-modeling,jian2024llmsgoodpragmaticspeakers}, while other studies examine scalar–adjective semantics \cite{lin-etal-2024-probing}, manner implicature \cite{cong2024manner}, and the resolution of non-literal intent in free-form generation \cite{DBLP:conf/acl/YerukolaVFS24}. 

\textbf{Training Phases of LLMs.}  
The typical pipeline for developing deployment-ready LLMs involves several sequential training phases. First, models are pre-trained on large-scale text corpora to acquire general-purpose language representations. This is followed by instruction tuning, where models are trained on curated input-output pairs to better follow human instructions \cite{mishra-etal-2022-cross, pmlr-v202-longpre23a}. We adopt the term “SFT” throughout this paper to align with current usage and emphasize its role as the first stage of alignment after pretraining. The final stage typically involves \textit{preference optimization}, commonly implemented via Proximal Policy Optimization \cite{schulman2017proximalpolicyoptimizationalgorithms} to align LLMs with human values. A recent and widely adopted PPO alternative: \textbf{Direct Preference Optimization (DPO)} simplifies PPO by avoiding reward modeling and policy optimization, instead directly optimizing model outputs to align with pairwise human preferences. Many open-source checkpoints e.g., OLMo-2 \cite{walsh2025} are released in DPO variants, which we adopt in our experimental comparisons.

A number of studies have investigated how these training stages affect downstream model behavior. For instance, \citet{song-etal-2025-dynamics} found that capabilities emerge at different rates during instruction tuning. \citet{kirk2024rlhf} conducted a systematic analysis of SFT and RLHF, reporting that RLHF improves out-of-distribution generalization but also reduces output diversity. Building on this line of work, we investigate these training phases in greater depth, with a particular focus on how each stage contributes to the emergence of pragmatic competence.

\section{\textsc{AltPrag}}

\begin{figure*}[!t]
    \centering
    \includegraphics[width=0.9\linewidth]{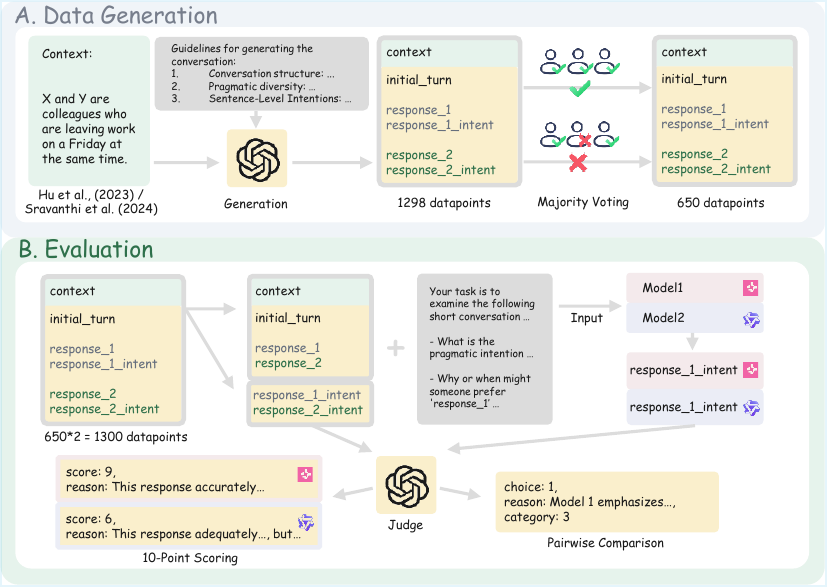}
    \caption{An illustration of the data generation process and evaluation workflow. After the majority voting phase, we construct a mirrored version by swapping the order of the two responses and their associated reference labels, resulting in a total of 1,300 data points.}
    \label{fig:teaser}
\end{figure*}

In pragmatics, alternatives refer to other plausible ways of expressing essentially the same semantic meaning. For a given prompt, the space of valid continuations is nearly infinite. Yet speakers routinely select one particular form over countless others—an act that often encodes subtle cues about their mental state, communicative intent, and contextual awareness \cite{degen2013alternatives}. For instance, as illustrated in \autoref{fig:alternative}, both responses plausibly continue the question “\textit{Are you going to the gym?}”, but they adopt distinct pragmatic stances. The first reply, “\textit{Yes, I’ve been slacking off lately and need to catch up},” offers a candid explanation, signaling openness and a willingness to connect through vulnerability. The second, “\textit{Do I ever do anything else on a Friday?}”, is more playful and rhetorically indirect, implying routine through sarcasm and suggesting a casual rapport between colleagues. Although both responses affirm the same propositional content—that the speaker is going to the gym—their divergent forms lead to different interpersonal effects, shaping how the speaker is perceived and how the utterance functions socially. Such variation across equally valid expressions highlights the contrastive nature of alternatives, making them especially well-suited for probing pragmatic competence in LLMs: the contrast between two semantically aligned yet pragmatically distinct replies creates a controlled setting for evaluating fine-grained reasoning about lexical form, intent, and context.

We thus leverage the concept of alternatives to build the \textsc{AltPrag} dataset, which probes LLMs’ sensitivity to speaker intent and social context. Each instance includes two replies with similar meaning but different pragmatic force, prompting the model to infer (1) the speaker’s underlying intention and (2) the circumstances motivating that particular wording. This dual task offers a practical lens on pragmatic competence, pushing the model to reason not just what is said, but why it is said that way in context, separating context-sensitive reasoning from simple semantic recall. We use GPT-4o \cite{openai2024gpt4ocard} to generate a reference set of alternative continuations with human-verified intent explanations (\autoref{fig:teaser}A).

\subsection{First-round Data Generation}
In the initial round of data generation, we build on the scenario-based dataset introduced by \citet{hu_fine-grained_2023} and the pragmatic benchmark proposed by \citet{sravanthi-etal-2024-pub}. For each data point, we extract the scenario description as contextual background and treat the target sentence as the root of a dialogue. Using this setup, we prompt GPT-4o \cite{openai2024gpt4ocard} to generate two contextually coherent but pragmatically distinct alternative continuations. The model is additionally instructed to provide explanations detailing the pragmatic functions conveyed by each alternative, as well as in what context a speaker would choose one over the other. This method allows us to elicit fine-grained pragmatic contrasts grounded in realistic and context-sensitive language use. Details on the prompt template and data postprocessing procedure are provided in \autoref{appendix:prompt}. In total, the first round of data generation yields 1298 datapoints.

\subsection{Human-in-the-loop Refinement}
Each datapoint was labeled as a \textit{pass} only if it met the evaluation criteria described below and was independently approved by all three annotators. Otherwise, it was marked as a \textit{fail}. Three authors with undergraduate training in pragmatics independently annotated each datapoint using the following criteria:
\vspace{-4pt}
\begin{itemize}[itemsep=0.35pt]
  \item[(1)] Both continuations must be coherent and contextually appropriate responses to the initial utterance.
  \item[(2)] Each natural language explanation must accurately capture the pragmatic function of its corresponding continuation and reflect nuanced speaker preferences.
\end{itemize}
\vspace{-4pt}
Out of the initial 1,298 raw examples generated by the model, 650 passed this filtering stage.

To augment the dataset for evaluation purposes, we apply a symmetric transformation: for each validated datapoint, we generate a mirrored version by swapping the order of the two responses and their corresponding explanations. This enables us to probe model judgments about each sentence independently. After augmentation, our final dataset contains 1,300 examples. A representative example datapoint is shown in Table~\ref{tab:example-datapoint}.

\begin{table}[h]
\centering
\small
\renewcommand{\arraystretch}{0.65}
\begin{tabularx}{\columnwidth}{lX}
\toprule
\textbf{Field} & \textbf{Content} \\
\midrule
\texttt{context} & X and Y are colleagues who are leaving work on a Friday at the same time. \\
\addlinespace
\texttt{initial\_turn} & \textit{Are you going to the gym?} \\
\addlinespace
\texttt{response\_1} & \textit{Yes, I've been slacking off lately and need to catch up.} \\
\addlinespace
\texttt{response\_1\_intent} & This response does not flout any maxim. It aims to directly acknowledge the intention to go to the gym while adding a self-reflective comment about recent habits. A speaker might prefer this reply over \texttt{response\_2} because it is straightforward, honest, and avoids sarcasm, making the interaction clearer and more sincere. This approach fits well if the speaker wishes to be open and relatable. \\
\addlinespace
\texttt{response\_2} & \textit{Do I ever do anything else on a Friday?} \\
\addlinespace
\texttt{response\_2\_intent} & This response flouts the maxim of Quality. It aims to use sarcasm to humorously imply that going to the gym on Fridays is routine for the speaker. A speaker might prefer this reply over \texttt{response\_1} because it injects playfulness and familiarity into the conversation, which can help maintain a lighthearted tone among colleagues. This style can strengthen rapport if both parties appreciate joking banter. \\
\bottomrule
\end{tabularx}
\caption{An example datapoint showing a complete conversation with two pragmatically distinct continuations and annotated intentions.}
\label{tab:example-datapoint}
\end{table}

\section{Experimental Setup}

In evaluations, we provide models with the conversations from \textsc{AltPrag} and prompt models to generate analogous explanations of pragmatic intent to our gold references, using these to evaluate models' pragmatic reasoning via 10-point scoring and pairwise comparison metrics (\autoref{fig:teaser}B).

\subsection{Evaluated LLM Variants}
To investigate how pragmatic competence develops across training stages, we evaluate a diverse set of open-source LLMs, covering different parameter scales and fine-tuning strategies:
\begin{itemize}[itemsep=0.35pt]
    \item \textit{OLMo-2 Series} \cite{walsh2025}: We evaluate OLMo-2 models at 7B, 13B, and 32B parameter scales. These models are trained on up to 6 trillion tokens and further refined using the Tülu 3 instruction-following and preference datasets.
    \item \textit{OLMoE-1B-7B} \cite{muennighoff2025olmoe}: This Mixture-of-Experts (MoE) model consists of 7 billion total parameters, with 1 billion active during inference. 
    \item \textit{LLaMA-3.1-Tülu-3 Series} \cite{grattafiori2024llama3herdmodels, lambert2025tulu}: Based on Meta's LLaMA-3.1 foundation models, we evaluate 8B and 70B parameter variants, each trained with the Tülu 3 post-training pipeline.
\end{itemize}
To further probe the emergence of pragmatic competence in base models, we additionally evaluate \textit{Qwen-3} base models at 0.6B, 1.7B, 4B, and 8B parameter sizes \cite{yang2025qwen3technicalreport}, as this project does not release instruction-tuned or preference-optimized checkpoints. This setup includes a broader comparison of baseline pragmatic abilities across model families and enables a fine-grained analysis of how pragmatic understanding emerges and evolves in LLMs, as well as the role of instruction tuning and preference optimization in shaping communicative competence.

\subsection{Prompting Strategy and Setup}
Evaluated models are prompted to generate an explanation of the pragmatic intention underlying each alternative in a conversational datapoint, as well as the pragmatic reasons why a speaker might prefer a given alternative to the other.

To mitigate the instability and underperformance commonly observed in interactions with base models, we adopt the \textsc{Urial} prompt template introduced by \citet{lin2024the}. This template is specifically designed to elicit more helpful and coherent outputs from base-stage LLMs without additional instruction tuning. For consistency and fairness across model stages, we apply the same template when evaluating SFT and DPO variants. The complete prompt template can be found in~\autoref{appendix:evaluation}.

To prevent evaluation inflation via format imitation, we adopt \textbf{zero-shot prompting} throughout, avoiding any in-prompt examples or structural cues. This ensures that models rely solely on their internal representations of pragmatic intent.

To control for variability, we fix decoding parameters across all runs: \texttt{max\_new\_tokens} = 256, \texttt{top\_k} = 50, \texttt{top\_p} = 1.0, and \texttt{temperature} = 0.5. Full configuration details appear in~\autoref{appendix:hyperparameters}.

\subsection{Evaluation Metrics}
\label{sec:eval-metrics}
We adopt two complementary LLM-as-a-Judge evaluation protocols \cite{DBLP:journals/corr/abs-2305-13711, fu-etal-2024-gptscore}, both employing GPT-4.1 \cite{openai2024gpt4ocard} as the evaluator to assess the quality of model-generated explanations of pragmatic intent.

\textbf{10-Point Scoring.} In this setting, the evaluator is provided with the conversation, reference intent explanation, and model-generated hypothesis intent explanation, and asked to assign each explanation a score on a 10-point scale, accompanied by a brief justification. This method allows for direct, fine-grained comparison of explanation quality across different model variants. The full prompt template is provided in ~\autoref{appendix:10-point-scoring}.

\textbf{Pairwise Win Rate.} To mitigate potential scoring biases and highlight relative differences across training stages, we also conduct pairwise comparisons between model variants (e.g., Base vs. SFT, SFT vs. DPO). For each pair, the evaluator is asked to determine which explanation better captures the speaker’s pragmatic intent. Drawing on the framework of pragmatic competence from \citet{doi:10.1177/21582440211011472}, we further instruct the evaluator to categorize each winning explanation into one of three dimensions:
\vspace{-4pt}
\begin{enumerate}[itemsep=0.35pt]
    \item \textit{Cognitive-pragmatic competence}: The explanation goes beyond literal meaning and identifies the speaker’s underlying communicative goal or intention.
    \item \textit{Pragmalinguistic competence}: The explanation highlights rhetorical strategies such as humor, irony, or self-deprecation and explains how these are used to manage interpersonal meaning.
    \item \textit{Sociopragmatic competence}: The explanation demonstrates awareness of social norms, roles, relationships, or context-sensitive appropriateness in the speaker’s choice.
\end{enumerate}
\vspace{-4pt}
Together, these two evaluation protocols enable both absolute assessment of explanation quality and nuanced, comparative analysis of pragmatic competence across training stages.

\section{Results}
\subsection{General Results}
We present our overall findings from both the 10-point scoring and pairwise win rate comparisons, focusing on how pragmatic competence develops across model training stages.

\begin{figure*}[!h]
    \centering
    \includegraphics[width=0.925\linewidth]{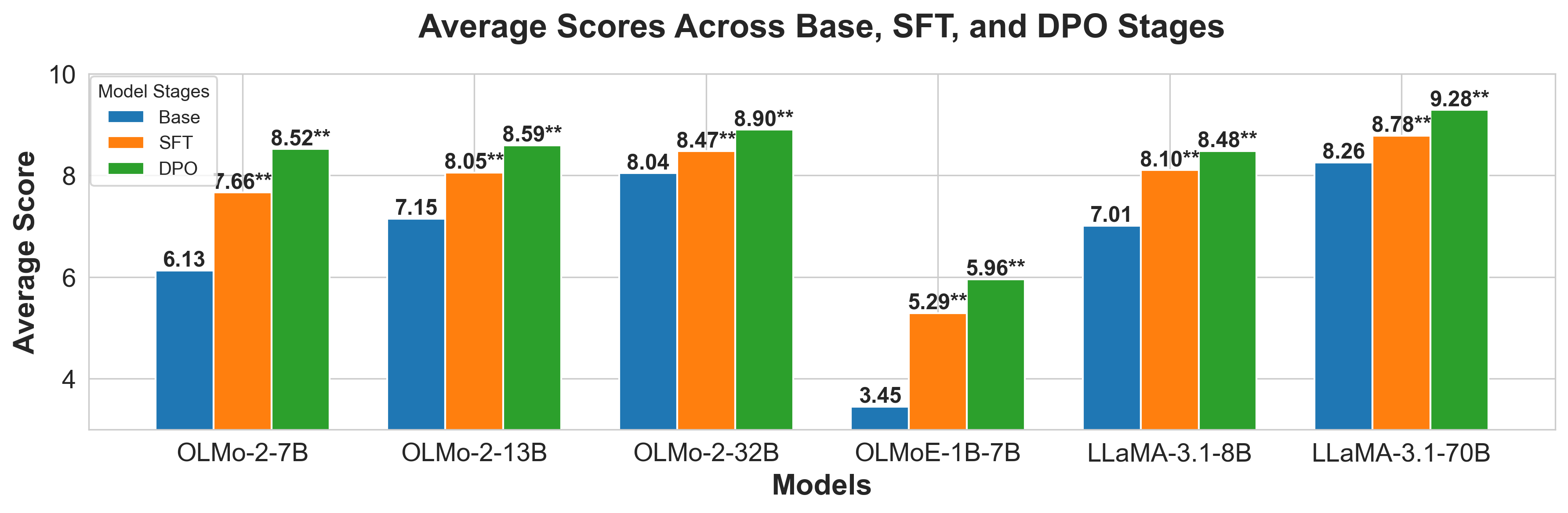}
    \caption{Average 10-point quality scores across Base, SFT, and DPO stages for different model families. Significance codes are based on Wilcoxon signed-rank tests comparing each stage with the previous one (e.g., SFT vs. Base, DPO vs. SFT). Asterisks denote statistical significance: * $p < 0.05$, ** $p < 0.01$. Base-stage results are not assigned significance codes as they are used as reference baselines.
}
    \label{fig:score-distribution}
\end{figure*}

\paragraph{10-Point Scoring.}
As shown in ~\autoref{fig:score-distribution}, models generally achieve higher scores as they progress from base to SFT to DPO stages. We conduct a Wilcoxon test \cite{wilcoxon1992individual} between the model and its immediately following training stage, and the results suggest that all score rises are statistically significant. First, we found that base models already demonstrate surprising competence, with average scores around 6 out of 10 for models with 7-8B parameters, indicating that early-stage models are already capable of non-trivial pragmatic inference without instruction tuning or preference optimization, likely benefiting from implicit exposure to pragmatic phenomena during large-scale pretraining. At the DPO stage, responses generally receive scores of 8 or higher, reflecting a marked alignment between model output and the intent conveyed in reference annotations. The complete model scores with distribution can be found in \autoref{appendix:scoring-distribution}. Example responses for each score range can be found in \autoref{appendix:scoring-results} for a better understanding of model performance.

\paragraph{Pairwise Win Rate.}
Consistent with the scoring results (\autoref{tab:pairwise-winrate-summary}), DPO models achieve the highest win rates in all head-to-head comparisons, followed by SFT and then base models. This pattern holds across model families and parameter scales, reinforcing the view that both SFT and DPO enhance pragmatic sensitivity. These findings support the idea that pragmatic competence emerges gradually, with measurable gains at each fine-tuning stage.

\begin{table}[!t]
\centering
\small
\setlength{\tabcolsep}{8pt} 
\renewcommand{\arraystretch}{0.95} 
\begin{tabularx}{\columnwidth}{c >{\hspace{0.15cm}}c<{\hspace{0.15cm}} c}
\toprule
\textbf{Model} & \textbf{Comparison} & \textbf{Win Rate (\%)} \\
\midrule
\texttt{OLMo-2-7B} & SFT over Base & 73.1 \\
\texttt{OLMo-2-7B} & DPO over SFT & 80.2 \\
\texttt{OLMo-2-7B} & DPO over Base & 89.2 \\
\texttt{OLMo-2-13B} & SFT over Base & 72.4 \\
\texttt{OLMo-2-13B} & DPO over SFT & 76.9 \\
\texttt{OLMo-2-13B} & DPO over Base & 90.1 \\
\texttt{OLMo-2-32B} & SFT over Base & 43.2 \\
\texttt{OLMo-2-32B} & DPO over SFT & 92.3 \\
\texttt{OLMo-2-32B} & DPO over Base & 85.8 \\
\texttt{OLMoE-1B-7B} & SFT over Base & 67.6 \\
\texttt{OLMoE-1B-7B} & DPO over SFT & 55.3 \\
\texttt{OLMoE-1B-7B} & DPO over Base & 80.3 \\
\texttt{Llama3-8B} & SFT over Base & 73.7 \\
\texttt{Llama3-8B} & DPO over SFT & 68.0 \\
\texttt{Llama3-8B} & DPO over Base & 83.6 \\
\texttt{Llama3-70B} & SFT over Base & 61.3 \\
\texttt{Llama3-70B} & DPO over SFT & 90.4 \\
\texttt{Llama3-70B} & DPO over Base & 89.0 \\
\bottomrule
\end{tabularx}
\caption{Pairwise win rate comparisons across model stages. Win rates are reported as the proportion of wins by the \textbf{later stage} model (e.g., SFT over Base).}
\label{tab:pairwise-winrate-summary}
\end{table}

\paragraph{Human and Model Agreements.}
To sanity–check the agreement between model judgments and human annotations, we collected human ratings on a subset of the data.\footnote{50 responses per stage (10-Point), yielding 150 items. 50 response pairs for each stage comparison (Pairwise), adding another 150 items. Two trained annotators rated independently; disagreements were resolved by adjudication.}  As~\autoref{tab:human_agreement}a shows, model scores correlate strongly with human evaluations in the 10-point Scoring task ($\rho\!\ge\!0.65$) and achieve substantial agreement in pairwise preference ($\kappa\!\ge\!0.56$), and ~\autoref{tab:human_agreement}b indicates overall reasonable inter-annotator reliability.

\begin{table}[h]
\centering
\small
\setlength{\tabcolsep}{8pt}
\renewcommand{\arraystretch}{0.95}

\begin{tabularx}{\columnwidth}{ccc}
\toprule
\multicolumn{3}{l}{\textbf{(a) Human--model agreement}} \\
\midrule
\textbf{Stages (Comparisons)} 
& \textbf{Spearman $\boldsymbol{\rho}$} 
& \textbf{Cohen $\boldsymbol{\kappa}$} \\
\midrule
Base (vs SFT) & 0.853 & 0.561 \\
SFT (vs DPO)  & 0.847 & 0.726 \\
DPO (vs Base) & 0.647 & 0.666 \\
\midrule
\multicolumn{3}{l}{\textbf{(b) Inter-annotator agreement (10-point scoring)}} \\
\midrule
\textbf{Stage} & \multicolumn{2}{c}{\textbf{Fleiss' $\boldsymbol{\kappa}$ (±1 tol.)}} \\
\midrule
Base & \multicolumn{2}{c}{0.766} \\
SFT  & \multicolumn{2}{c}{0.528} \\
DPO  & \multicolumn{2}{c}{0.399} \\
\bottomrule
\end{tabularx}

\caption{Agreement between model judgments and human annotations on the subset of \textsc{AltPrag}.
Fleiss' $\kappa$ is computed with a $\pm1$ tolerance due to the fine-grained 10-point scale, i.e., two ratings are treated as consistent if they differ by at most one point. All correlations are significant ($p<0.01$).}
\label{tab:human_agreement}
\end{table}

\subsection{Does Pragmatic Competence Scale?}
We further analyze how pragmatic competence scales with two key factors: model size and pretraining data volume.

\paragraph{Scaling with Model Size.}
We observe that larger models tend to achieve better pragmatic competence across families. This trend holds across evaluated model families, including OLMo-2 \cite{walsh2025} and LLaMA-3.1-Tülu-3 \cite{grattafiori2024llama3herdmodels, lambert2025tulu}. While OLMo-2 models show improvements across sizes, each scale is trained on different amounts of pretraining data, making it difficult to attribute gains solely to model size. To isolate the effect of scaling, we compare LLaMA-3.1 7B and 70B models trained on the same corpus: the 70B model achieves a much higher win rate (66\% vs. 34\%), indicating that increased capacity enhances pragmatic competence. A similar trend is observed in Qwen-3 models \cite{yang2025qwen3technicalreport}, which vary in size but share the same pretraining data—larger models consistently outperform smaller ones.

\begin{figure} [h]
    \centering
    \includegraphics[width=0.75\columnwidth]{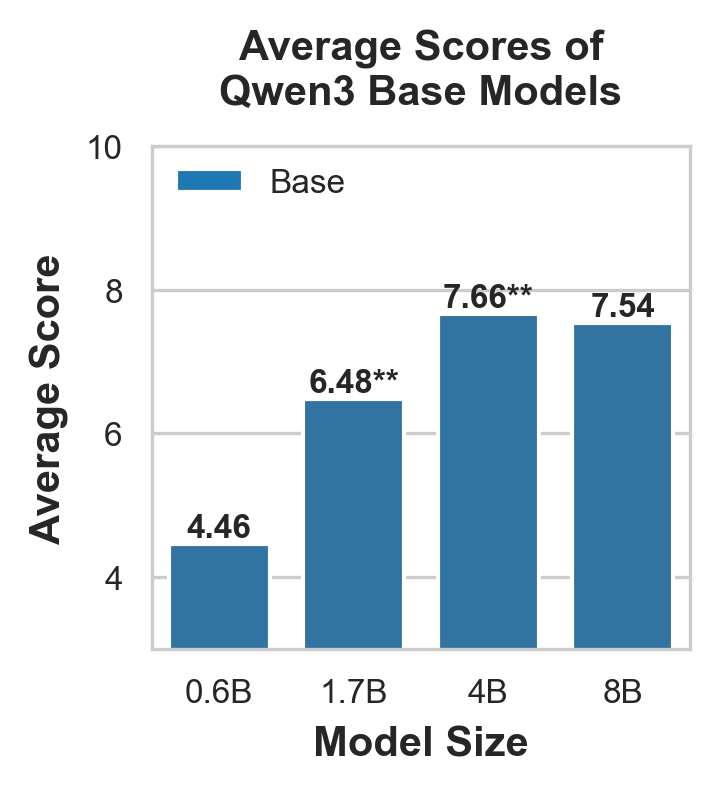}
    \caption{The Qwen-3 series achieves comparatively higher scores with fewer parameters, illustrating that scaling pretraining data size can enhance a model's capacity for pragmatic reasoning.}
    \label{fig:qwen3}
\end{figure}

\begin{figure*}[t]
    \centering
    \begin{subfigure}[b]{0.22\textwidth}
        \centering
        \includegraphics[width=\textwidth]{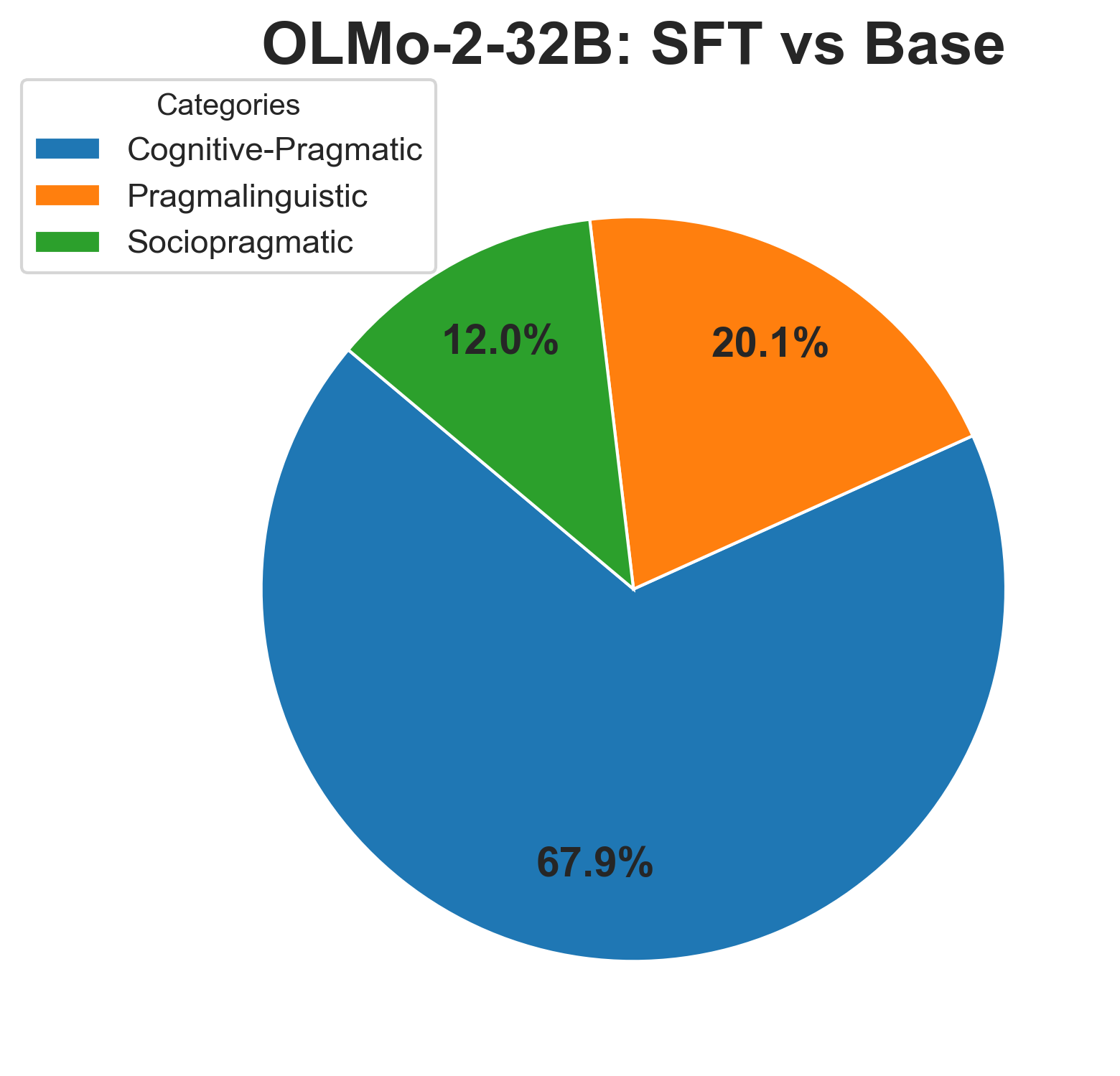}
        \caption{OLMo-2-32B SFT vs Base}
    \end{subfigure}
    \hfill
    \begin{subfigure}[b]{0.22\textwidth}
        \centering
        \includegraphics[width=\textwidth]{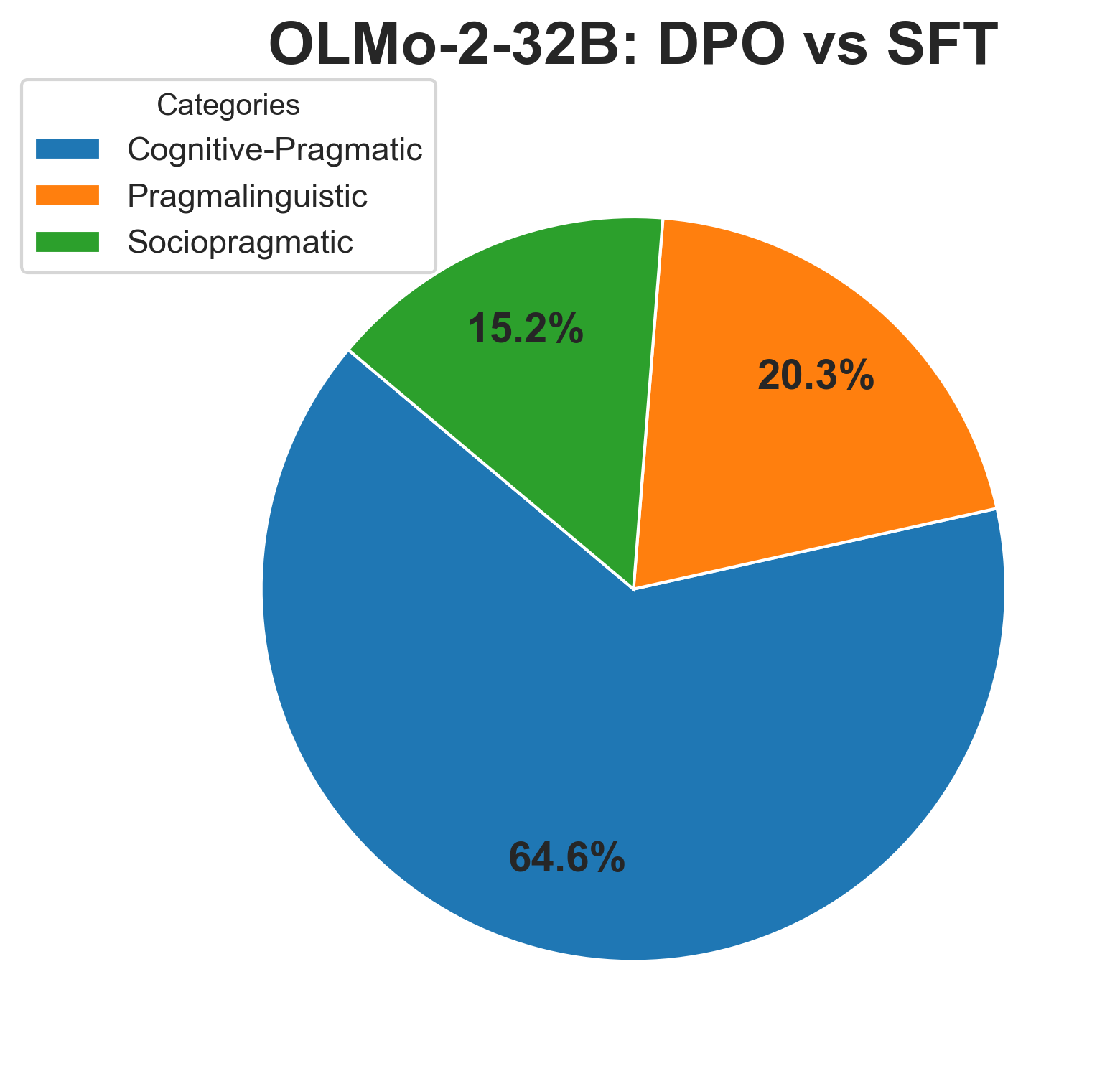}
        \caption{OLMo-2-32B DPO vs SFT}
    \end{subfigure}
    \hfill
    \begin{subfigure}[b]{0.22\textwidth}
        \centering
        \includegraphics[width=\textwidth]{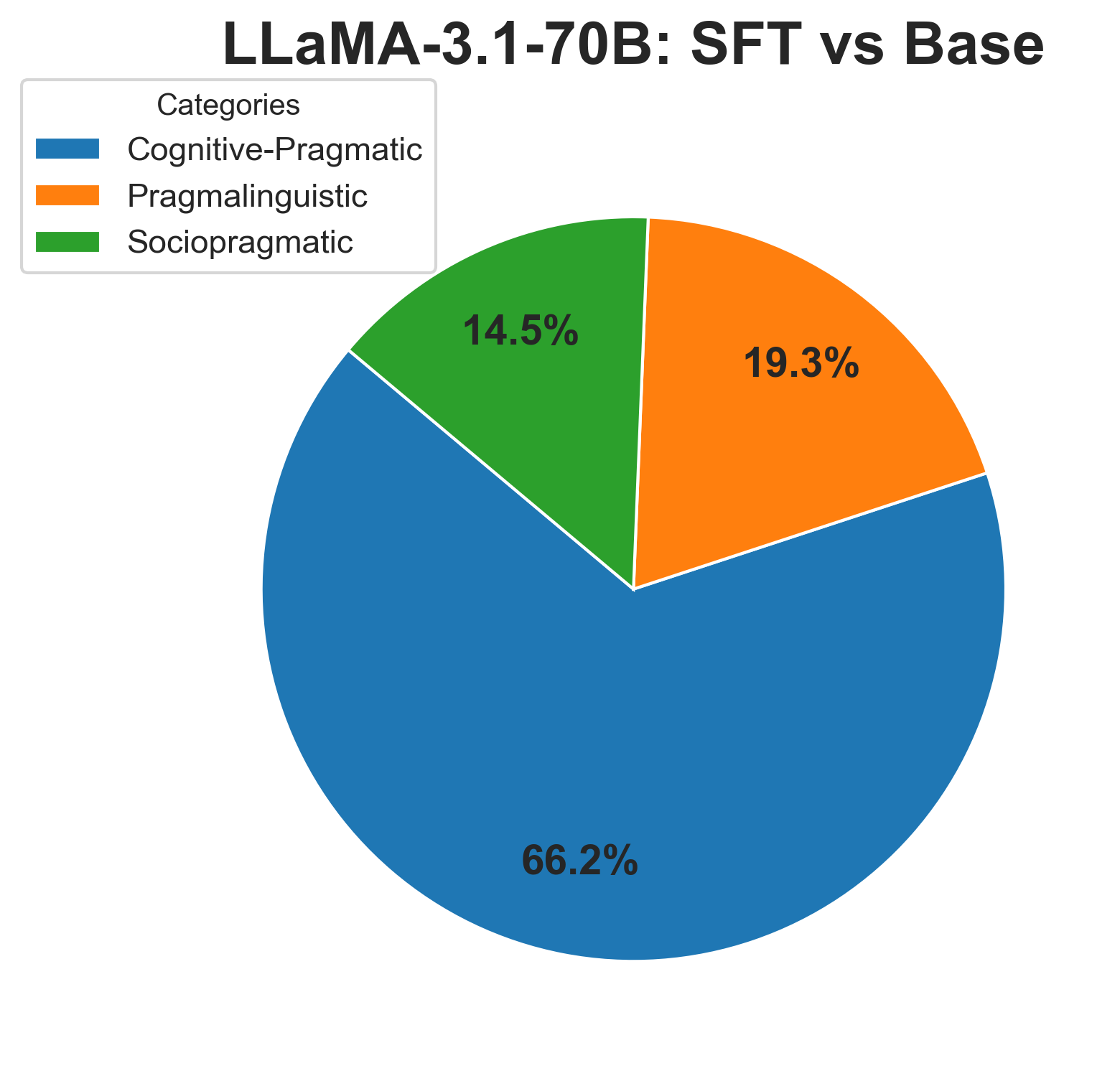}
        \caption{LLaMA-3.1-70B SFT vs Base}
    \end{subfigure}
    \hfill
    \begin{subfigure}[b]{0.22\textwidth}
        \centering
        \includegraphics[width=\textwidth]{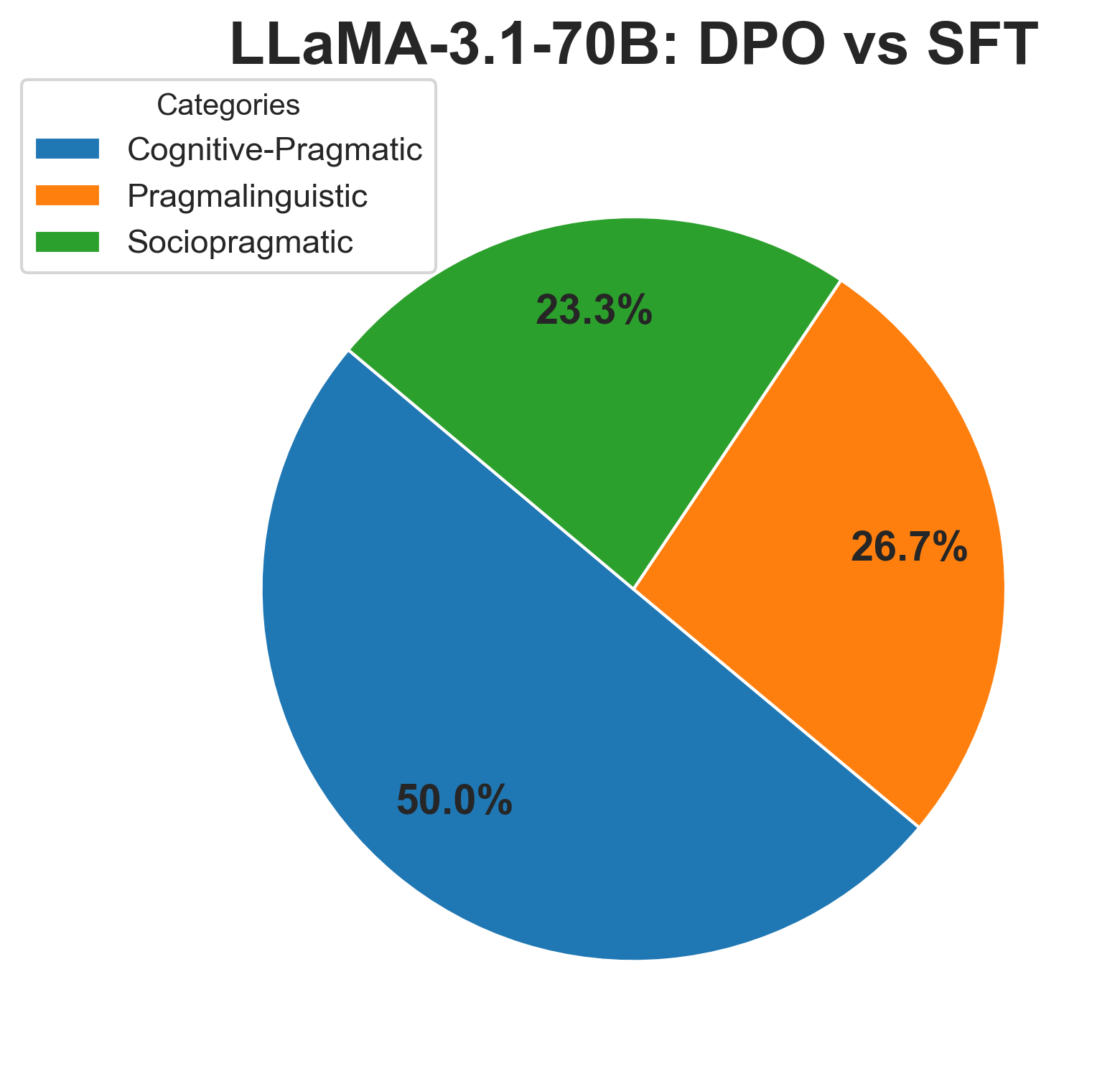}
        \caption{LLaMA-3.1-70B DPO vs SFT}
    \end{subfigure}
    
    \caption{Distribution of winning explanation categories across selected model comparisons. While both SFT and DPO stages are dominated by cognitive-pragmatic explanations, the DPO stage shows a notable increase in sociopragmatic responses, indicating enhanced sensitivity to social context and appropriateness.}
    \label{fig:end-to-end-cat-distribution}
\end{figure*}

\paragraph{Scaling with Pretraining Data.}
We also find that pretraining data volume contributes significantly to base model performance. The Qwen-3 series, trained on 36T tokens \cite{yang2025qwen3technicalreport}, shows relatively strong pragmatic competence across parameter scales, performing better than other models of similar size trained on smaller corpora—such as the OLMo-2 series, which is trained on 4T/5T/6T tokens for the 7B/13B/32B models, respectively \cite{walsh2025}. Notably, the Qwen-3 1.7B model achieves a higher average score (6.48) than the OLMo-2 7B model (6.13), illustrating how pretraining scale alone can improve models' ability to infer pragmatic intent, suggesting that larger pretraining corpora can also contribute to enhancing pragmatic abilities.

Taken together, our results show that both parameter and pretraining corpus size shape pragmatic ability. While larger models tend to perform better, our findings highlight the often underappreciated role of pretraining data quality and scale—particularly in the emergence of early-stage pragmatic competence. These results underscore the need to consider pretraining data as an important factor in shaping pragmatic abilities.

\subsection{Where Do SFT and DPO Help?}
To better understand the role of fine-tuning and preference optimization in shaping pragmatic competence, in \autoref{fig:end-to-end-cat-distribution} we visualize the distribution of winning explanations across the three aforementioned categories of pragmatic competence from \citet{doi:10.1177/21582440211011472}: cognitive-pragmatic, pragmalinguistic, and sociopragmatic.

We find that \textit{cognitive-pragmatic competence}, the ability to go beyond literal meaning and infer the speaker’s communicative goal, is the primary justification for wins across all model stages. This trend is especially pronounced in SFT stage, where cognitive-pragmatic explanations account for the majority of wins over base models. In the OLMo-2-32B SFT variant, 66.7\% of winning explanations fall into this category, suggesting that supervised fine-tuning primarily strengthens the model’s ability to capture intended meaning.

While cognitive-pragmatic competence remains dominant in DPO, we observe a continued strengthening of this ability compared to SFT stage, indicating that DPO further refines models' understanding of speaker intent. In parallel, we also observe a shift toward more \textit{sociopragmatic competence}—the ability to recognize social roles, politeness strategies, and contextual appropriateness—suggesting that the DPO stage broadens the scope of pragmatic strategies beyond purely cognitive interpretations. A complete case study with analysis can be found in \autoref{appendix:case-study} and the full comparison results can be found in \autoref{appendix:end-to-end-prag}.

\section{Discussion}
In this paper, we revisit the findings presented in \citet{ruis} and utilize the concept of alternative to construct a dataset for evaluating LLMs at various training stages. Our results provide a complement to the previous findings: although instruction tuning and DPO surely help, base models already show non-trivial pragmatic competence in contrastive pragmatic reasoning tasks.

\paragraph{The Role of Pretraining.}
Our findings underscore the foundational role of pretraining in shaping LLMs’ pragmatic competence. Even base models exhibit notable sensitivity to speaker intent and context: Qwen-3 series, despite having smaller parameter counts to its counterparts, performs competitively across pragmatic tasks. Notably, Qwen-3 models are trained on 36T tokens, one of the largest reported pretraining corpora, suggesting that the scale and quality of pretraining data can significantly enhance pragmatic reasoning, independent of model size. This also aligns with \citet{yue2025does, ai2025rethinkingreflectionpretraining}, who report that base models already demonstrate strong reasoning capabilities. Extending their findings to the pragmatic domain, we argue that much of an LLM’s pragmatic ability is rooted in pretraining, reinforcing its importance not only for general reasoning but also for socially competent language use.

\paragraph{Revisiting Goldilocks: Improved Base Model Performance.}
To further contextualize our findings and make a fair comparison, we replicate the experiments proposed in \citet{ruis} using the zero-shot setting and find that modern base models substantially outperform those in the original study, even surpassing GPT-3-175B. As shown in \autoref{tab:goldilocks-replication}, larger OLMo-2 base models reach accuracy levels above 70\%, highlighting the increased pragmatic competence brought by better quality of pretraining without instruction tuning. These results also exhibit clear scaling patterns, with accuracy improving as model size increases, consistently scales with model size, highlighting the role of high-quality pretraining in fostering pragmatic competence. 
\vspace{-2pt}
\begin{table}[h]
\centering
\small
\renewcommand{\arraystretch}{0.95} 
\begin{tabularx}{\columnwidth}{c >{\centering\arraybackslash}X}
\toprule
\textbf{Model} & \textbf{Accuracy (\%)} \\
\midrule
GPT-2-xl (Goldilocks)     & 51.3 \\
OPT-13B (Goldilocks)      & 61.0 \\
GPT-3-175B (Goldilocks) & 57.7 \\
\midrule
OLMo-2-7B            & 71.7 \\
OLMo-2-13B           & 70.5 \\
OLMo-2-32B           & 75.5 \\
Qwen3-0.6B-Base           & 63.8 \\
Qwen3-1.7B-Base           & 67.3 \\
Qwen3-4B-Base             & 69.5 \\
\bottomrule
\end{tabularx}
\caption{Accuracy on the Goldilocks implicature reasoning task. Top section shows original results reported by \citet{ruis}; bottom section reports our own evaluation of modern base models using the same experimental setup.}
\label{tab:goldilocks-replication}
\end{table}
\vspace{-4pt}

\paragraph{Beyond Pretraining: Refining Pragmatic Competence through SFT and DPO.}
While pre-training provides the basic substrate for pragmatic reasoning, our experiments show that supervised fine-tuning (SFT) markedly strengthens a model’s cognitive-pragmatic competence by improving inference of speaker intent. DPO further enhances sociopragmatic abilities by improving the model’s sensitivity to social context, roles, and politeness norms. \citet{wu-etal-2024-rethinking} argue that preference optimization may offer a "near-free lunch," improving pragmatic ability without degrading general performance. Our findings reinforce this view, highlighting the critical role of preference optimization in advancing pragmatic competence, especially in socially grounded interpretations.

\section*{Limitations}
This work faces two primary limitations. First, the current dataset does not explicitly distinguish among different types of pragmatic phenomena, such as humor, indirect speech, or irony, which limits our ability to analyze how models at various training stages handle specific subcategories of pragmatics. Second, while we include multiple model families and architectures, all evaluated models across training stages are developed by the same organization (AI2). This shared provenance may introduce systematic biases, potentially limiting the generalizability of our findings. Finally, while \textsc{AltPrag} enables fine-grained analysis through controlled contrasts, its moderate scale constrains coverage across the full space of pragmatic phenomena, motivating future work on broader and more diverse datasets.

\bibliography{custom}

@inproceedings{ruis,
author = {Ruis, Laura and Khan, Akbir and Biderman, Stella and Hooker, Sara and Rockt\"{a}schel, Tim and Grefenstette, Edward},
title = {The goldilocks of pragmatic understanding: fine-tuning strategy matters for implicature resolution by LLMs},
year = {2023},
publisher = {Curran Associates Inc.},
address = {Red Hook, NY, USA},
booktitle = {Proceedings of the 37th International Conference on Neural Information Processing Systems},
articleno = {913},
numpages = {79},
location = {New Orleans, LA, USA},
series = {NIPS '23}
}

@incollection{sadock1978testing,
  title={On testing for conversational implicature},
  author={Sadock, Jerrold M},
  booktitle={Pragmatics},
  pages={281--297},
  year={1978},
  publisher={Brill}
}

@article{karttunen1974presupposition,
  title={Presupposition and linguistic context},
  author={Karttunen, Lauri},
  year={1974},
  publisher={Walter de Gruyter, Berlin/New York Berlin, New York}
}

@incollection{searle1975indirect,
  title={Indirect speech acts},
  author={Searle, John R},
  booktitle={Speech acts},
  pages={59--82},
  year={1975},
  publisher={Brill}
}

@inproceedings{hu_fine-grained_2023,
    title = {A fine-grained comparison of pragmatic language understanding in humans and language models},
    url = {https://aclanthology.org/2023.acl-long.230/},
    code = {https://github.com/jennhu/lm-pragmatics},
    author = {Hu, Jennifer and Floyd, Sammy and Jouravlev, Olessia and Fedorenko, Evelina and Gibson, Edward},
    year = {2023},
    booktitle = {Proceedings of the 61st Annual Meeting of the Association for Computational Linguistics}
}

@article{
doi:10.1073/pnas.2405460121,
author = {Michal Kosinski },
title = {Evaluating large language models in theory of mind tasks},
journal = {Proceedings of the National Academy of Sciences},
volume = {121},
number = {45},
pages = {e2405460121},
year = {2024},
doi = {10.1073/pnas.2405460121},
URL = {https://www.pnas.org/doi/abs/10.1073/pnas.2405460121},
eprint = {https://www.pnas.org/doi/pdf/10.1073/pnas.2405460121}}

@inproceedings{shapira-etal-2024-clever,
    title = "Clever Hans or Neural Theory of Mind? Stress Testing Social Reasoning in Large Language Models",
    author = "Shapira, Natalie  and
      Levy, Mosh  and
      Alavi, Seyed Hossein  and
      Zhou, Xuhui  and
      Choi, Yejin  and
      Goldberg, Yoav  and
      Sap, Maarten  and
      Shwartz, Vered",
    editor = "Graham, Yvette  and
      Purver, Matthew",
    booktitle = "Proceedings of the 18th Conference of the European Chapter of the Association for Computational Linguistics (Volume 1: Long Papers)",
    month = mar,
    year = "2024",
    address = "St. Julian{'}s, Malta",
    publisher = "Association for Computational Linguistics",
    url = "https://aclanthology.org/2024.eacl-long.138/",
    pages = "2257--2273"
}

@inproceedings{chen-etal-2024-tombench,
    title = "{T}o{MB}ench: Benchmarking Theory of Mind in Large Language Models",
    author = "Chen, Zhuang  and
      Wu, Jincenzi  and
      Zhou, Jinfeng  and
      Wen, Bosi  and
      Bi, Guanqun  and
      Jiang, Gongyao  and
      Cao, Yaru  and
      Hu, Mengting  and
      Lai, Yunghwei  and
      Xiong, Zexuan  and
      Huang, Minlie",
    editor = "Ku, Lun-Wei  and
      Martins, Andre  and
      Srikumar, Vivek",
    booktitle = "Proceedings of the 62nd Annual Meeting of the Association for Computational Linguistics (Volume 1: Long Papers)",
    month = aug,
    year = "2024",
    address = "Bangkok, Thailand",
    publisher = "Association for Computational Linguistics",
    url = "https://aclanthology.org/2024.acl-long.847/",
    doi = "10.18653/v1/2024.acl-long.847",
    pages = "15959--15983"
}

@inproceedings{sravanthi-etal-2024-pub,
    title = "{PUB}: A Pragmatics Understanding Benchmark for Assessing {LLM}s' Pragmatics Capabilities",
    author = "Sravanthi, Settaluri  and
      Doshi, Meet  and
      Tankala, Pavan  and
      Murthy, Rudra  and
      Dabre, Raj  and
      Bhattacharyya, Pushpak",
    editor = "Ku, Lun-Wei  and
      Martins, Andre  and
      Srikumar, Vivek",
    booktitle = "Findings of the Association for Computational Linguistics: ACL 2024",
    month = aug,
    year = "2024",
    address = "Bangkok, Thailand",
    publisher = "Association for Computational Linguistics",
    url = "https://aclanthology.org/2024.findings-acl.719/",
    doi = "10.18653/v1/2024.findings-acl.719",
    pages = "12075--12097"
}

@misc{jones2025largelanguagemodelspass,
      title={Large Language Models Pass the Turing Test}, 
      author={Cameron R. Jones and Benjamin K. Bergen},
      year={2025},
      eprint={2503.23674},
      archivePrefix={arXiv},
      primaryClass={cs.CL},
      url={https://arxiv.org/abs/2503.23674}, 
}

@inproceedings{wu-etal-2024-rethinking,
    title = "Rethinking Pragmatics in Large Language Models: Towards Open-Ended Evaluation and Preference Tuning",
    author = "Wu, Shengguang  and
      Yang, Shusheng  and
      Chen, Zhenglun  and
      Su, Qi",
    editor = "Al-Onaizan, Yaser  and
      Bansal, Mohit  and
      Chen, Yun-Nung",
    booktitle = "Proceedings of the 2024 Conference on Empirical Methods in Natural Language Processing",
    month = nov,
    year = "2024",
    address = "Miami, Florida, USA",
    publisher = "Association for Computational Linguistics",
    url = "https://aclanthology.org/2024.emnlp-main.1258/",
    doi = "10.18653/v1/2024.emnlp-main.1258",
    pages = "22583--22599"
}

@article{GEORGE20202316,
title = {Conversational implicatures in English dialogue: Annotated dataset},
journal = {Procedia Computer Science},
volume = {171},
pages = {2316-2323},
year = {2020},
note = {Third International Conference on Computing and Network Communications (CoCoNet'19)},
issn = {1877-0509},
doi = {https://doi.org/10.1016/j.procs.2020.04.251},
url = {https://www.sciencedirect.com/science/article/pii/S1877050920312436},
author = {Elizabeth Jasmi George and Radhika Mamidi}
}

@article{cong2024manner,
  title={Manner implicatures in large language models},
  author={Cong, Yan},
  journal={Scientific Reports},
  volume={14},
  number={1},
  pages={29113},
  year={2024},
  publisher={Nature Publishing Group UK London}
}

@inproceedings{shaikh-etal-2023-modeling,
    title = "Modeling Cross-Cultural Pragmatic Inference with Codenames Duet",
    author = "Shaikh, Omar  and
      Ziems, Caleb  and
      Held, William  and
      Pariani, Aryan  and
      Morstatter, Fred  and
      Yang, Diyi",
    editor = "Rogers, Anna  and
      Boyd-Graber, Jordan  and
      Okazaki, Naoaki",
    booktitle = "Findings of the Association for Computational Linguistics: ACL 2023",
    month = jul,
    year = "2023",
    address = "Toronto, Canada",
    publisher = "Association for Computational Linguistics",
    url = "https://aclanthology.org/2023.findings-acl.410/",
    doi = "10.18653/v1/2023.findings-acl.410",
    pages = "6550--6569"
}

@misc{jian2024llmsgoodpragmaticspeakers,
      title={Are LLMs good pragmatic speakers?}, 
      author={Mingyue Jian and N. Siddharth},
      year={2024},
      eprint={2411.01562},
      archivePrefix={arXiv},
      primaryClass={cs.CL},
      url={https://arxiv.org/abs/2411.01562}, 
}

@InProceedings{pmlr-v202-longpre23a,
  title = 	 {The Flan Collection: Designing Data and Methods for Effective Instruction Tuning},
  author =       {Longpre, Shayne and Hou, Le and Vu, Tu and Webson, Albert and Chung, Hyung Won and Tay, Yi and Zhou, Denny and Le, Quoc V and Zoph, Barret and Wei, Jason and Roberts, Adam},
  booktitle = 	 {Proceedings of the 40th International Conference on Machine Learning},
  pages = 	 {22631--22648},
  year = 	 {2023},
  editor = 	 {Krause, Andreas and Brunskill, Emma and Cho, Kyunghyun and Engelhardt, Barbara and Sabato, Sivan and Scarlett, Jonathan},
  volume = 	 {202},
  series = 	 {Proceedings of Machine Learning Research},
  month = 	 {23--29 Jul},
  publisher =    {PMLR},
  url = 	 {https://proceedings.mlr.press/v202/longpre23a.html}
}

@inproceedings{mishra-etal-2022-cross,
    title = "Cross-Task Generalization via Natural Language Crowdsourcing Instructions",
    author = "Mishra, Swaroop  and
      Khashabi, Daniel  and
      Baral, Chitta  and
      Hajishirzi, Hannaneh",
    editor = "Muresan, Smaranda  and
      Nakov, Preslav  and
      Villavicencio, Aline",
    booktitle = "Proceedings of the 60th Annual Meeting of the Association for Computational Linguistics (Volume 1: Long Papers)",
    month = may,
    year = "2022",
    address = "Dublin, Ireland",
    publisher = "Association for Computational Linguistics",
    url = "https://aclanthology.org/2022.acl-long.244/",
    doi = "10.18653/v1/2022.acl-long.244",
    pages = "3470--3487"
}

@inproceedings{song-etal-2025-dynamics,
    title = "Dynamics of Instruction Fine-Tuning for {C}hinese Large Language Models",
    author = "Song, Chiyu  and
      Zhou, Zhanchao  and
      Yan, Jianhao  and
      Fei, Yuejiao  and
      Lan, Zhenzhong  and
      Zhang, Yue",
    editor = "Rambow, Owen  and
      Wanner, Leo  and
      Apidianaki, Marianna  and
      Al-Khalifa, Hend  and
      Eugenio, Barbara Di  and
      Schockaert, Steven",
    booktitle = "Proceedings of the 31st International Conference on Computational Linguistics",
    month = jan,
    year = "2025",
    address = "Abu Dhabi, UAE",
    publisher = "Association for Computational Linguistics",
    url = "https://aclanthology.org/2025.coling-main.689/",
    pages = "10345--10366"
}

@inproceedings{kirk2024rlhf,
  title        = {Understanding the Effects of {RLHF} on {LLM} Generalisation and Diversity},
  author       = {Kirk, Robert and Mediratta, Ishita and Nalmpantis, Christoforos and Luketina, Jelena and Hambro, Eric and Grefenstette, Edward and Raileanu, Roberta},
  booktitle    = {International Conference on Learning Representations (ICLR)},
  year         = {2024},
  url          = {https://openreview.net/forum?id=PXD3FAVHJT},
}

@book{degen2013alternatives,
  title={Alternatives in pragmatic reasoning},
  author={Degen, Judith},
  year={2013},
  publisher={University of Rochester}
}

@misc{openai2024gpt4ocard,
      title={GPT-4o System Card}, 
      author={OpenAI and : and Aaron Hurst and Adam Lerer and Adam P. Goucher and Adam Perelman and Aditya Ramesh and Aidan Clark and AJ Ostrow and Akila Welihinda and Alan Hayes and Alec Radford and Aleksander Mądry and Alex Baker-Whitcomb and Alex Beutel and Alex Borzunov and Alex Carney and Alex Chow and Alex Kirillov and Alex Nichol and Alex Paino and Alex Renzin and Alex Tachard Passos and Alexander Kirillov and Alexi Christakis and Alexis Conneau and Ali Kamali and Allan Jabri and Allison Moyer and Allison Tam and Amadou Crookes and Amin Tootoochian and Amin Tootoonchian and Ananya Kumar and Andrea Vallone and Andrej Karpathy and Andrew Braunstein and Andrew Cann and Andrew Codispoti and Andrew Galu and Andrew Kondrich and Andrew Tulloch and Andrey Mishchenko and Angela Baek and Angela Jiang and Antoine Pelisse and Antonia Woodford and Anuj Gosalia and Arka Dhar and Ashley Pantuliano and Avi Nayak and Avital Oliver and Barret Zoph and Behrooz Ghorbani and Ben Leimberger and Ben Rossen and Ben Sokolowsky and Ben Wang and Benjamin Zweig and Beth Hoover and Blake Samic and Bob McGrew and Bobby Spero and Bogo Giertler and Bowen Cheng and Brad Lightcap and Brandon Walkin and Brendan Quinn and Brian Guarraci and Brian Hsu and Bright Kellogg and Brydon Eastman and Camillo Lugaresi and Carroll Wainwright and Cary Bassin and Cary Hudson and Casey Chu and Chad Nelson and Chak Li and Chan Jun Shern and Channing Conger and Charlotte Barette and Chelsea Voss and Chen Ding and Cheng Lu and Chong Zhang and Chris Beaumont and Chris Hallacy and Chris Koch and Christian Gibson and Christina Kim and Christine Choi and Christine McLeavey and Christopher Hesse and Claudia Fischer and Clemens Winter and Coley Czarnecki and Colin Jarvis and Colin Wei and Constantin Koumouzelis and Dane Sherburn and Daniel Kappler and Daniel Levin and Daniel Levy and David Carr and David Farhi and David Mely and David Robinson and David Sasaki and Denny Jin and Dev Valladares and Dimitris Tsipras and Doug Li and Duc Phong Nguyen and Duncan Findlay and Edede Oiwoh and Edmund Wong and Ehsan Asdar and Elizabeth Proehl and Elizabeth Yang and Eric Antonow and Eric Kramer and Eric Peterson and Eric Sigler and Eric Wallace and Eugene Brevdo and Evan Mays and Farzad Khorasani and Felipe Petroski Such and Filippo Raso and Francis Zhang and Fred von Lohmann and Freddie Sulit and Gabriel Goh and Gene Oden and Geoff Salmon and Giulio Starace and Greg Brockman and Hadi Salman and Haiming Bao and Haitang Hu and Hannah Wong and Haoyu Wang and Heather Schmidt and Heather Whitney and Heewoo Jun and Hendrik Kirchner and Henrique Ponde de Oliveira Pinto and Hongyu Ren and Huiwen Chang and Hyung Won Chung and Ian Kivlichan and Ian O'Connell and Ian O'Connell and Ian Osband and Ian Silber and Ian Sohl and Ibrahim Okuyucu and Ikai Lan and Ilya Kostrikov and Ilya Sutskever and Ingmar Kanitscheider and Ishaan Gulrajani and Jacob Coxon and Jacob Menick and Jakub Pachocki and James Aung and James Betker and James Crooks and James Lennon and Jamie Kiros and Jan Leike and Jane Park and Jason Kwon and Jason Phang and Jason Teplitz and Jason Wei and Jason Wolfe and Jay Chen and Jeff Harris and Jenia Varavva and Jessica Gan Lee and Jessica Shieh and Ji Lin and Jiahui Yu and Jiayi Weng and Jie Tang and Jieqi Yu and Joanne Jang and Joaquin Quinonero Candela and Joe Beutler and Joe Landers and Joel Parish and Johannes Heidecke and John Schulman and Jonathan Lachman and Jonathan McKay and Jonathan Uesato and Jonathan Ward and Jong Wook Kim and Joost Huizinga and Jordan Sitkin and Jos Kraaijeveld and Josh Gross and Josh Kaplan and Josh Snyder and Joshua Achiam and Joy Jiao and Joyce Lee and Juntang Zhuang and Justyn Harriman and Kai Fricke and Kai Hayashi and Karan Singhal and Katy Shi and Kavin Karthik and Kayla Wood and Kendra Rimbach and Kenny Hsu and Kenny Nguyen and Keren Gu-Lemberg and Kevin Button and Kevin Liu and Kiel Howe and Krithika Muthukumar and Kyle Luther and Lama Ahmad and Larry Kai and Lauren Itow and Lauren Workman and Leher Pathak and Leo Chen and Li Jing and Lia Guy and Liam Fedus and Liang Zhou and Lien Mamitsuka and Lilian Weng and Lindsay McCallum and Lindsey Held and Long Ouyang and Louis Feuvrier and Lu Zhang and Lukas Kondraciuk and Lukasz Kaiser and Luke Hewitt and Luke Metz and Lyric Doshi and Mada Aflak and Maddie Simens and Madelaine Boyd and Madeleine Thompson and Marat Dukhan and Mark Chen and Mark Gray and Mark Hudnall and Marvin Zhang and Marwan Aljubeh and Mateusz Litwin and Matthew Zeng and Max Johnson and Maya Shetty and Mayank Gupta and Meghan Shah and Mehmet Yatbaz and Meng Jia Yang and Mengchao Zhong and Mia Glaese and Mianna Chen and Michael Janner and Michael Lampe and Michael Petrov and Michael Wu and Michele Wang and Michelle Fradin and Michelle Pokrass and Miguel Castro and Miguel Oom Temudo de Castro and Mikhail Pavlov and Miles Brundage and Miles Wang and Minal Khan and Mira Murati and Mo Bavarian and Molly Lin and Murat Yesildal and Nacho Soto and Natalia Gimelshein and Natalie Cone and Natalie Staudacher and Natalie Summers and Natan LaFontaine and Neil Chowdhury and Nick Ryder and Nick Stathas and Nick Turley and Nik Tezak and Niko Felix and Nithanth Kudige and Nitish Keskar and Noah Deutsch and Noel Bundick and Nora Puckett and Ofir Nachum and Ola Okelola and Oleg Boiko and Oleg Murk and Oliver Jaffe and Olivia Watkins and Olivier Godement and Owen Campbell-Moore and Patrick Chao and Paul McMillan and Pavel Belov and Peng Su and Peter Bak and Peter Bakkum and Peter Deng and Peter Dolan and Peter Hoeschele and Peter Welinder and Phil Tillet and Philip Pronin and Philippe Tillet and Prafulla Dhariwal and Qiming Yuan and Rachel Dias and Rachel Lim and Rahul Arora and Rajan Troll and Randall Lin and Rapha Gontijo Lopes and Raul Puri and Reah Miyara and Reimar Leike and Renaud Gaubert and Reza Zamani and Ricky Wang and Rob Donnelly and Rob Honsby and Rocky Smith and Rohan Sahai and Rohit Ramchandani and Romain Huet and Rory Carmichael and Rowan Zellers and Roy Chen and Ruby Chen and Ruslan Nigmatullin and Ryan Cheu and Saachi Jain and Sam Altman and Sam Schoenholz and Sam Toizer and Samuel Miserendino and Sandhini Agarwal and Sara Culver and Scott Ethersmith and Scott Gray and Sean Grove and Sean Metzger and Shamez Hermani and Shantanu Jain and Shengjia Zhao and Sherwin Wu and Shino Jomoto and Shirong Wu and Shuaiqi and Xia and Sonia Phene and Spencer Papay and Srinivas Narayanan and Steve Coffey and Steve Lee and Stewart Hall and Suchir Balaji and Tal Broda and Tal Stramer and Tao Xu and Tarun Gogineni and Taya Christianson and Ted Sanders and Tejal Patwardhan and Thomas Cunninghman and Thomas Degry and Thomas Dimson and Thomas Raoux and Thomas Shadwell and Tianhao Zheng and Todd Underwood and Todor Markov and Toki Sherbakov and Tom Rubin and Tom Stasi and Tomer Kaftan and Tristan Heywood and Troy Peterson and Tyce Walters and Tyna Eloundou and Valerie Qi and Veit Moeller and Vinnie Monaco and Vishal Kuo and Vlad Fomenko and Wayne Chang and Weiyi Zheng and Wenda Zhou and Wesam Manassra and Will Sheu and Wojciech Zaremba and Yash Patil and Yilei Qian and Yongjik Kim and Youlong Cheng and Yu Zhang and Yuchen He and Yuchen Zhang and Yujia Jin and Yunxing Dai and Yury Malkov},
      year={2024},
      eprint={2410.21276},
      archivePrefix={arXiv},
      primaryClass={cs.CL},
      url={https://arxiv.org/abs/2410.21276}, 
}

@inproceedings{
walsh2025,
title={2 {OLM}o 2 Furious ({COLM}{\textquoteright}s Version)},
author={Evan Pete Walsh and Luca Soldaini and Dirk Groeneveld and Kyle Lo and Shane Arora and Akshita Bhagia and Yuling Gu and Shengyi Huang and Matt Jordan and Nathan Lambert and Dustin Schwenk and Oyvind Tafjord and Taira Anderson and David Atkinson and Faeze Brahman and Christopher Clark and Pradeep Dasigi and Nouha Dziri and Allyson Ettinger and Michal Guerquin and David Heineman and Hamish Ivison and Pang Wei Koh and Jiacheng Liu and Saumya Malik and William Merrill and Lester James Validad Miranda and Jacob Morrison and Tyler Murray and Crystal Nam and Jake Poznanski and Valentina Pyatkin and Aman Rangapur and Michael Schmitz and Sam Skjonsberg and David Wadden and Christopher Wilhelm and Michael Wilson and Luke Zettlemoyer and Ali Farhadi and Noah A. Smith and Hannaneh Hajishirzi},
booktitle={Second Conference on Language Modeling},
year={2025},
url={https://openreview.net/forum?id=2ezugTT9kU}
}

@inproceedings{
lambert2025tulu,
title={Tulu 3: Pushing Frontiers in Open Language Model Post-Training},
author={Nathan Lambert and Jacob Morrison and Valentina Pyatkin and Shengyi Huang and Hamish Ivison and Faeze Brahman and Lester James Validad Miranda and Alisa Liu and Nouha Dziri and Xinxi Lyu and Yuling Gu and Saumya Malik and Victoria Graf and Jena D. Hwang and Jiangjiang Yang and Ronan Le Bras and Oyvind Tafjord and Christopher Wilhelm and Luca Soldaini and Noah A. Smith and Yizhong Wang and Pradeep Dasigi and Hannaneh Hajishirzi},
booktitle={Second Conference on Language Modeling},
year={2025},
url={https://openreview.net/forum?id=i1uGbfHHpH}
}

@inproceedings{
muennighoff2025olmoe,
title={{OLM}oE: Open Mixture-of-Experts Language Models},
author={Niklas Muennighoff and Luca Soldaini and Dirk Groeneveld and Kyle Lo and Jacob Morrison and Sewon Min and Weijia Shi and Evan Pete Walsh and Oyvind Tafjord and Nathan Lambert and Yuling Gu and Shane Arora and Akshita Bhagia and Dustin Schwenk and David Wadden and Alexander Wettig and Binyuan Hui and Tim Dettmers and Douwe Kiela and Ali Farhadi and Noah A. Smith and Pang Wei Koh and Amanpreet Singh and Hannaneh Hajishirzi},
booktitle={The Thirteenth International Conference on Learning Representations},
year={2025},
url={https://openreview.net/forum?id=xXTkbTBmqq}
}

@misc{grattafiori2024llama3herdmodels,
      title={The Llama 3 Herd of Models}, 
      author={Aaron Grattafiori and Abhimanyu Dubey and Abhinav Jauhri and Abhinav Pandey and Abhishek Kadian and Ahmad Al-Dahle and Aiesha Letman and Akhil Mathur and Alan Schelten and Alex Vaughan and Amy Yang and Angela Fan and Anirudh Goyal and Anthony Hartshorn and Aobo Yang and Archi Mitra and Archie Sravankumar and Artem Korenev and Arthur Hinsvark and Arun Rao and Aston Zhang and Aurelien Rodriguez and Austen Gregerson and Ava Spataru and Baptiste Roziere and Bethany Biron and Binh Tang and Bobbie Chern and Charlotte Caucheteux and Chaya Nayak and Chloe Bi and Chris Marra and Chris McConnell and Christian Keller and Christophe Touret and Chunyang Wu and Corinne Wong and Cristian Canton Ferrer and Cyrus Nikolaidis and Damien Allonsius and Daniel Song and Danielle Pintz and Danny Livshits and Danny Wyatt and David Esiobu and Dhruv Choudhary and Dhruv Mahajan and Diego Garcia-Olano and Diego Perino and Dieuwke Hupkes and Egor Lakomkin and Ehab AlBadawy and Elina Lobanova and Emily Dinan and Eric Michael Smith and Filip Radenovic and Francisco Guzmán and Frank Zhang and Gabriel Synnaeve and Gabrielle Lee and Georgia Lewis Anderson and Govind Thattai and Graeme Nail and Gregoire Mialon and Guan Pang and Guillem Cucurell and Hailey Nguyen and Hannah Korevaar and Hu Xu and Hugo Touvron and Iliyan Zarov and Imanol Arrieta Ibarra and Isabel Kloumann and Ishan Misra and Ivan Evtimov and Jack Zhang and Jade Copet and Jaewon Lee and Jan Geffert and Jana Vranes and Jason Park and Jay Mahadeokar and Jeet Shah and Jelmer van der Linde and Jennifer Billock and Jenny Hong and Jenya Lee and Jeremy Fu and Jianfeng Chi and Jianyu Huang and Jiawen Liu and Jie Wang and Jiecao Yu and Joanna Bitton and Joe Spisak and Jongsoo Park and Joseph Rocca and Joshua Johnstun and Joshua Saxe and Junteng Jia and Kalyan Vasuden Alwala and Karthik Prasad and Kartikeya Upasani and Kate Plawiak and Ke Li and Kenneth Heafield and Kevin Stone and Khalid El-Arini and Krithika Iyer and Kshitiz Malik and Kuenley Chiu and Kunal Bhalla and Kushal Lakhotia and Lauren Rantala-Yeary and Laurens van der Maaten and Lawrence Chen and Liang Tan and Liz Jenkins and Louis Martin and Lovish Madaan and Lubo Malo and Lukas Blecher and Lukas Landzaat and Luke de Oliveira and Madeline Muzzi and Mahesh Pasupuleti and Mannat Singh and Manohar Paluri and Marcin Kardas and Maria Tsimpoukelli and Mathew Oldham and Mathieu Rita and Maya Pavlova and Melanie Kambadur and Mike Lewis and Min Si and Mitesh Kumar Singh and Mona Hassan and Naman Goyal and Narjes Torabi and Nikolay Bashlykov and Nikolay Bogoychev and Niladri Chatterji and Ning Zhang and Olivier Duchenne and Onur Çelebi and Patrick Alrassy and Pengchuan Zhang and Pengwei Li and Petar Vasic and Peter Weng and Prajjwal Bhargava and Pratik Dubal and Praveen Krishnan and Punit Singh Koura and Puxin Xu and Qing He and Qingxiao Dong and Ragavan Srinivasan and Raj Ganapathy and Ramon Calderer and Ricardo Silveira Cabral and Robert Stojnic and Roberta Raileanu and Rohan Maheswari and Rohit Girdhar and Rohit Patel and Romain Sauvestre and Ronnie Polidoro and Roshan Sumbaly and Ross Taylor and Ruan Silva and Rui Hou and Rui Wang and Saghar Hosseini and Sahana Chennabasappa and Sanjay Singh and Sean Bell and Seohyun Sonia Kim and Sergey Edunov and Shaoliang Nie and Sharan Narang and Sharath Raparthy and Sheng Shen and Shengye Wan and Shruti Bhosale and Shun Zhang and Simon Vandenhende and Soumya Batra and Spencer Whitman and Sten Sootla and Stephane Collot and Suchin Gururangan and Sydney Borodinsky and Tamar Herman and Tara Fowler and Tarek Sheasha and Thomas Georgiou and Thomas Scialom and Tobias Speckbacher and Todor Mihaylov and Tong Xiao and Ujjwal Karn and Vedanuj Goswami and Vibhor Gupta and Vignesh Ramanathan and Viktor Kerkez and Vincent Gonguet and Virginie Do and Vish Vogeti and Vítor Albiero and Vladan Petrovic and Weiwei Chu and Wenhan Xiong and Wenyin Fu and Whitney Meers and Xavier Martinet and Xiaodong Wang and Xiaofang Wang and Xiaoqing Ellen Tan and Xide Xia and Xinfeng Xie and Xuchao Jia and Xuewei Wang and Yaelle Goldschlag and Yashesh Gaur and Yasmine Babaei and Yi Wen and Yiwen Song and Yuchen Zhang and Yue Li and Yuning Mao and Zacharie Delpierre Coudert and Zheng Yan and Zhengxing Chen and Zoe Papakipos and Aaditya Singh and Aayushi Srivastava and Abha Jain and Adam Kelsey and Adam Shajnfeld and Adithya Gangidi and Adolfo Victoria and Ahuva Goldstand and Ajay Menon and Ajay Sharma and Alex Boesenberg and Alexei Baevski and Allie Feinstein and Amanda Kallet and Amit Sangani and Amos Teo and Anam Yunus and Andrei Lupu and Andres Alvarado and Andrew Caples and Andrew Gu and Andrew Ho and Andrew Poulton and Andrew Ryan and Ankit Ramchandani and Annie Dong and Annie Franco and Anuj Goyal and Aparajita Saraf and Arkabandhu Chowdhury and Ashley Gabriel and Ashwin Bharambe and Assaf Eisenman and Azadeh Yazdan and Beau James and Ben Maurer and Benjamin Leonhardi and Bernie Huang and Beth Loyd and Beto De Paola and Bhargavi Paranjape and Bing Liu and Bo Wu and Boyu Ni and Braden Hancock and Bram Wasti and Brandon Spence and Brani Stojkovic and Brian Gamido and Britt Montalvo and Carl Parker and Carly Burton and Catalina Mejia and Ce Liu and Changhan Wang and Changkyu Kim and Chao Zhou and Chester Hu and Ching-Hsiang Chu and Chris Cai and Chris Tindal and Christoph Feichtenhofer and Cynthia Gao and Damon Civin and Dana Beaty and Daniel Kreymer and Daniel Li and David Adkins and David Xu and Davide Testuggine and Delia David and Devi Parikh and Diana Liskovich and Didem Foss and Dingkang Wang and Duc Le and Dustin Holland and Edward Dowling and Eissa Jamil and Elaine Montgomery and Eleonora Presani and Emily Hahn and Emily Wood and Eric-Tuan Le and Erik Brinkman and Esteban Arcaute and Evan Dunbar and Evan Smothers and Fei Sun and Felix Kreuk and Feng Tian and Filippos Kokkinos and Firat Ozgenel and Francesco Caggioni and Frank Kanayet and Frank Seide and Gabriela Medina Florez and Gabriella Schwarz and Gada Badeer and Georgia Swee and Gil Halpern and Grant Herman and Grigory Sizov and Guangyi and Zhang and Guna Lakshminarayanan and Hakan Inan and Hamid Shojanazeri and Han Zou and Hannah Wang and Hanwen Zha and Haroun Habeeb and Harrison Rudolph and Helen Suk and Henry Aspegren and Hunter Goldman and Hongyuan Zhan and Ibrahim Damlaj and Igor Molybog and Igor Tufanov and Ilias Leontiadis and Irina-Elena Veliche and Itai Gat and Jake Weissman and James Geboski and James Kohli and Janice Lam and Japhet Asher and Jean-Baptiste Gaya and Jeff Marcus and Jeff Tang and Jennifer Chan and Jenny Zhen and Jeremy Reizenstein and Jeremy Teboul and Jessica Zhong and Jian Jin and Jingyi Yang and Joe Cummings and Jon Carvill and Jon Shepard and Jonathan McPhie and Jonathan Torres and Josh Ginsburg and Junjie Wang and Kai Wu and Kam Hou U and Karan Saxena and Kartikay Khandelwal and Katayoun Zand and Kathy Matosich and Kaushik Veeraraghavan and Kelly Michelena and Keqian Li and Kiran Jagadeesh and Kun Huang and Kunal Chawla and Kyle Huang and Lailin Chen and Lakshya Garg and Lavender A and Leandro Silva and Lee Bell and Lei Zhang and Liangpeng Guo and Licheng Yu and Liron Moshkovich and Luca Wehrstedt and Madian Khabsa and Manav Avalani and Manish Bhatt and Martynas Mankus and Matan Hasson and Matthew Lennie and Matthias Reso and Maxim Groshev and Maxim Naumov and Maya Lathi and Meghan Keneally and Miao Liu and Michael L. Seltzer and Michal Valko and Michelle Restrepo and Mihir Patel and Mik Vyatskov and Mikayel Samvelyan and Mike Clark and Mike Macey and Mike Wang and Miquel Jubert Hermoso and Mo Metanat and Mohammad Rastegari and Munish Bansal and Nandhini Santhanam and Natascha Parks and Natasha White and Navyata Bawa and Nayan Singhal and Nick Egebo and Nicolas Usunier and Nikhil Mehta and Nikolay Pavlovich Laptev and Ning Dong and Norman Cheng and Oleg Chernoguz and Olivia Hart and Omkar Salpekar and Ozlem Kalinli and Parkin Kent and Parth Parekh and Paul Saab and Pavan Balaji and Pedro Rittner and Philip Bontrager and Pierre Roux and Piotr Dollar and Polina Zvyagina and Prashant Ratanchandani and Pritish Yuvraj and Qian Liang and Rachad Alao and Rachel Rodriguez and Rafi Ayub and Raghotham Murthy and Raghu Nayani and Rahul Mitra and Rangaprabhu Parthasarathy and Raymond Li and Rebekkah Hogan and Robin Battey and Rocky Wang and Russ Howes and Ruty Rinott and Sachin Mehta and Sachin Siby and Sai Jayesh Bondu and Samyak Datta and Sara Chugh and Sara Hunt and Sargun Dhillon and Sasha Sidorov and Satadru Pan and Saurabh Mahajan and Saurabh Verma and Seiji Yamamoto and Sharadh Ramaswamy and Shaun Lindsay and Shaun Lindsay and Sheng Feng and Shenghao Lin and Shengxin Cindy Zha and Shishir Patil and Shiva Shankar and Shuqiang Zhang and Shuqiang Zhang and Sinong Wang and Sneha Agarwal and Soji Sajuyigbe and Soumith Chintala and Stephanie Max and Stephen Chen and Steve Kehoe and Steve Satterfield and Sudarshan Govindaprasad and Sumit Gupta and Summer Deng and Sungmin Cho and Sunny Virk and Suraj Subramanian and Sy Choudhury and Sydney Goldman and Tal Remez and Tamar Glaser and Tamara Best and Thilo Koehler and Thomas Robinson and Tianhe Li and Tianjun Zhang and Tim Matthews and Timothy Chou and Tzook Shaked and Varun Vontimitta and Victoria Ajayi and Victoria Montanez and Vijai Mohan and Vinay Satish Kumar and Vishal Mangla and Vlad Ionescu and Vlad Poenaru and Vlad Tiberiu Mihailescu and Vladimir Ivanov and Wei Li and Wenchen Wang and Wenwen Jiang and Wes Bouaziz and Will Constable and Xiaocheng Tang and Xiaojian Wu and Xiaolan Wang and Xilun Wu and Xinbo Gao and Yaniv Kleinman and Yanjun Chen and Ye Hu and Ye Jia and Ye Qi and Yenda Li and Yilin Zhang and Ying Zhang and Yossi Adi and Youngjin Nam and Yu and Wang and Yu Zhao and Yuchen Hao and Yundi Qian and Yunlu Li and Yuzi He and Zach Rait and Zachary DeVito and Zef Rosnbrick and Zhaoduo Wen and Zhenyu Yang and Zhiwei Zhao and Zhiyu Ma},
      year={2024},
      eprint={2407.21783},
      archivePrefix={arXiv},
      primaryClass={cs.AI},
      url={https://arxiv.org/abs/2407.21783}, 
}

@inproceedings{
lin2024the,
title={The Unlocking Spell on Base {LLM}s:  Rethinking Alignment via In-Context Learning},
author={Bill Yuchen Lin and Abhilasha Ravichander and Ximing Lu and Nouha Dziri and Melanie Sclar and Khyathi Chandu and Chandra Bhagavatula and Yejin Choi},
booktitle={The Twelfth International Conference on Learning Representations},
year={2024},
url={https://openreview.net/forum?id=wxJ0eXwwda}
}

@article{DBLP:journals/corr/abs-2305-13711,
  publtype={informal},
  author={Yen-Ting Lin and Yun-Nung Chen},
  title={LLM-Eval: Unified Multi-Dimensional Automatic Evaluation for Open-Domain Conversations with Large Language Models},
  year={2023},
  cdate={1672531200000},
  journal={CoRR},
  volume={abs/2305.13711},
  url={https://doi.org/10.48550/arXiv.2305.13711}
}

@inproceedings{fu-etal-2024-gptscore,
    title = "{GPTS}core: Evaluate as You Desire",
    author = "Fu, Jinlan  and
      Ng, See-Kiong  and
      Jiang, Zhengbao  and
      Liu, Pengfei",
    editor = "Duh, Kevin  and
      Gomez, Helena  and
      Bethard, Steven",
    booktitle = "Proceedings of the 2024 Conference of the North American Chapter of the Association for Computational Linguistics: Human Language Technologies (Volume 1: Long Papers)",
    month = jun,
    year = "2024",
    address = "Mexico City, Mexico",
    publisher = "Association for Computational Linguistics",
    url = "https://aclanthology.org/2024.naacl-long.365/",
    doi = "10.18653/v1/2024.naacl-long.365",
    pages = "6556--6576"
}

@article{doi:10.1177/21582440211011472,
author = {Tiaoyuan Mao and Shanhua He},
title ={An Integrated Approach to Pragmatic Competence: Its Framework and Properties},

journal = {SAGE Open},
volume = {11},
number = {2},
pages = {21582440211011472},
year = {2021},
doi = {10.1177/21582440211011472},

URL = { 
    
        https://doi.org/10.1177/21582440211011472
    
    

},
eprint = { 
    
        https://doi.org/10.1177/21582440211011472
    
    

}
}

@misc{schulman2017proximalpolicyoptimizationalgorithms,
      title={Proximal Policy Optimization Algorithms}, 
      author={John Schulman and Filip Wolski and Prafulla Dhariwal and Alec Radford and Oleg Klimov},
      year={2017},
      eprint={1707.06347},
      archivePrefix={arXiv},
      primaryClass={cs.LG},
      url={https://arxiv.org/abs/1707.06347}, 
}

@misc{yang2025qwen3technicalreport,
      title={Qwen3 Technical Report}, 
      author={An Yang and Anfeng Li and Baosong Yang and Beichen Zhang and Binyuan Hui and Bo Zheng and Bowen Yu and Chang Gao and Chengen Huang and Chenxu Lv and Chujie Zheng and Dayiheng Liu and Fan Zhou and Fei Huang and Feng Hu and Hao Ge and Haoran Wei and Huan Lin and Jialong Tang and Jian Yang and Jianhong Tu and Jianwei Zhang and Jianxin Yang and Jiaxi Yang and Jing Zhou and Jingren Zhou and Junyang Lin and Kai Dang and Keqin Bao and Kexin Yang and Le Yu and Lianghao Deng and Mei Li and Mingfeng Xue and Mingze Li and Pei Zhang and Peng Wang and Qin Zhu and Rui Men and Ruize Gao and Shixuan Liu and Shuang Luo and Tianhao Li and Tianyi Tang and Wenbiao Yin and Xingzhang Ren and Xinyu Wang and Xinyu Zhang and Xuancheng Ren and Yang Fan and Yang Su and Yichang Zhang and Yinger Zhang and Yu Wan and Yuqiong Liu and Zekun Wang and Zeyu Cui and Zhenru Zhang and Zhipeng Zhou and Zihan Qiu},
      year={2025},
      eprint={2505.09388},
      archivePrefix={arXiv},
      primaryClass={cs.CL},
      url={https://arxiv.org/abs/2505.09388}, 
}

@inproceedings{
yue2025does,
title={Does Reinforcement Learning Really Incentivize Reasoning Capacity in {LLM}s Beyond the Base Model?},
author={Yang Yue and Zhiqi Chen and Rui Lu and Andrew Zhao and Zhaokai Wang and Yang Yue and Shiji Song and Gao Huang},
booktitle={2nd AI for Math Workshop @ ICML 2025},
year={2025},
url={https://openreview.net/forum?id=upehLVgq1b}
}

@misc{ai2025rethinkingreflectionpretraining,
      title={Rethinking Reflection in Pre-Training}, 
      author={{Essential AI} and : and Darsh J Shah and Peter Rushton and Somanshu Singla and Mohit Parmar and Kurt Smith and Yash Vanjani and Ashish Vaswani and Adarsh Chaluvaraju and Andrew Hojel and Andrew Ma and Anil Thomas and Anthony Polloreno and Ashish Tanwer and Burhan Drak Sibai and Divya S Mansingka and Divya Shivaprasad and Ishaan Shah and Karl Stratos and Khoi Nguyen and Michael Callahan and Michael Pust and Mrinal Iyer and Philip Monk and Platon Mazarakis and Ritvik Kapila and Saurabh Srivastava and Tim Romanski},
      year={2025},
      eprint={2504.04022},
      archivePrefix={arXiv},
      primaryClass={cs.CL},
      url={https://arxiv.org/abs/2504.04022}, 
}

@incollection{wilcoxon1992individual,
  title={Individual comparisons by ranking methods},
  author={Wilcoxon, Frank},
  booktitle={Breakthroughs in statistics: Methodology and distribution},
  pages={196--202},
  year={1992},
  publisher={Springer}
}

@inproceedings{lin-etal-2024-probing,
    title = "Probing Large Language Models for Scalar Adjective Lexical Semantics and Scalar Diversity Pragmatics",
    author = "Lin, Fangru  and
      Altshuler, Daniel  and
      Pierrehumbert, Janet B.",
    editor = "Calzolari, Nicoletta  and
      Kan, Min-Yen  and
      Hoste, Veronique  and
      Lenci, Alessandro  and
      Sakti, Sakriani  and
      Xue, Nianwen",
    booktitle = "Proceedings of the 2024 Joint International Conference on Computational Linguistics, Language Resources and Evaluation (LREC-COLING 2024)",
    month = may,
    year = "2024",
    address = "Torino, Italia",
    publisher = "ELRA and ICCL",
    url = "https://preview.aclanthology.org/fix-sig-urls/2024.lrec-main.1141/",
    pages = "13033--13049"
}

@article{DBLP:journals/corr/abs-2502-12378,
  publtype={informal},
  author={Bolei Ma and Yuting Li and Wei Zhou and Ziwei Gong and Yang Janet Liu and Katja Jasinskaja and Annemarie Friedrich and Julia Hirschberg and Frauke Kreuter and Barbara Plank},
  title={Pragmatics in the Era of Large Language Models: A Survey on Datasets, Evaluation, Opportunities and Challenges},
  year={2025},
  month={February},
  cdate={1738368000000},
  journal={CoRR},
  volume={abs/2502.12378},
  url={https://doi.org/10.48550/arXiv.2502.12378}
}

@inproceedings{DBLP:conf/acl/YerukolaVFS24,
  author={Akhila Yerukola and Saujas Vaduguru and Daniel Fried and Maarten Sap},
  title={Is the Pope Catholic? Yes, the Pope is Catholic. Generative Evaluation of Non-Literal Intent Resolution in LLMs},
  year={2024},
  cdate={1704067200000},
  pages={265-275},
  url={https://aclanthology.org/2024.acl-short.26},
  booktitle={ACL (Short Papers)},
}

\appendix

\section{Prompt Template for Data Generation}
\label{appendix:prompt}

We use a detailed prompt to instruct GPT-4o during the initial data generation phase. This prompt guides the model to construct a tree-structured dialogue rooted in a given scenario. For each scenario, we ask the model to generate \textbf{three} semantically coherent but pragmatically distinct alternative replies to augment our dataset to form \textbf{three one-to-one pairs} as datapoints (e.g. response A/B, B/C, and A/C). It also requests natural language justifications that explain each reply’s pragmatic function and its potential conversational effect. For all evaluation tasks, we use four H100 GPU. We obtain all open-source models from HuggingFace. The full prompt used in generation is shown below.

\begin{tcolorbox}[promptbox, title=Prompt Template for Data Generation, colback=white, colframe=black!75]
\small

\textbf{Task Overview:}  
In this task, we will explore \textbf{pragmatic alternatives} in conversation by constructing a multi-round, \textbf{tree-structured} dialogue. The goal is to analyze why a speaker might choose one alternative over another based on \textbf{pragmatic effects} rather than purely semantic differences.

\medskip

\textbf{Guidelines for Generating the Conversation:}

\textbf{1. Conversation Structure (Tree)} 

- The conversation will expand in a \textbf{1-3} structure (4 sentences total).  

- The second turn introduces \textbf{three alternative responses}, each reflecting a \textbf{distinct pragmatic intention}.  

- The conversation alternates speakers:  

  - Root: A initiates the conversation.  
  
  - Second layer: B responds with three alternatives (B1, B2, B3).  

\medskip

\textbf{2. Pragmatic Diversity (Not Constrained to Specific Effects)}  

- Each pair of responses must be \textbf{semantically similar but pragmatically different}.  

- Responses should vary across a \textbf{broad range of pragmatic effects}, such as:  

  - \textbf{Politeness \& Indirectness} (*e.g.,* "Could you open the window?" vs. "Open the window.")  
  
  - \textbf{Emphasis \& Focus} (*e.g.,* "I think you did great." vs. "You did great, no doubt about it.")  
  
  - \textbf{Formality \& Register} (*e.g.,* "He passed away." vs. "He died.")  
  
  - \textbf{Tone \& Emotion} (*e.g.,* "I’m not sure." vs. "I don’t think so.")  
  
  - \textbf{Hedging \& Certainty} (*e.g.,* "It might work." vs. "This will definitely work.")  
  
  - \textbf{Conversational Strategies} (*e.g.,* "That’s a good idea!" vs. "I see where you’re coming from.")  
  
- \textbf{Avoid overly simplistic or directly opposing responses} (e.g., "Yes." vs. "No.").  

- \textbf{Make sure the conversation not being too neutral, be drastic if necessary.}  

- \textbf{Strictly avoid repeating the context or simple agreement.} E.g., if context says A, avoid saying “yes, it’s A”. Be natural!  

\medskip

\textbf{3. Violation of Gricean Maxims for Each Alternative}  

Sometimes, speakers intentionally violate these maxims to imply something else (implicature). For example:  

- \textbf{Flouting Quantity:} Giving too much or too little information to imply something (e.g., "Some students passed the exam" implies not all did).  

- \textbf{Flouting Quality:} Using sarcasm or irony (e.g., "Oh, great! Another surprise quiz!" when unhappy).  

- \textbf{Flouting Relation:} Responding indirectly to suggest something (e.g., A: "Did you finish the report?" B: "I had a really busy weekend.").  

- \textbf{Flouting Manner:} Being intentionally vague (e.g., "Let's just say things didn't go as planned.").  

- \textbf{Ensure at least 2 of the 3 alternatives contain a violation of Gricean Maxims.}  

\medskip

\textbf{4. Sentence-Level Intentions}  

Each response must explicitly state its \textbf{pragmatic intention} in the JSON output:  

- \textbf{GM:} Which Gricean Maxim type does the response flout. If none, type “None”.  

- \textbf{Intention:} Explanation including:  

  - How the three alternatives differ pragmatically. 
  
  - When or why a speaker might choose each alternative. 
  
  - The effect each alternative has on the conversation (*e.g.,* softens tone, strengthens commitment, redirects focus).  

\medskip

\textbf{5. Demonstration}  

Below is an example of the JSON output:

Context: X and Y are colleagues who are leaving work on a Friday at the same time.  

Question: Do you want me to give you a lift home?

Root: Do you want me to give you a lift home?

B1: Oh, that would be great! I was dreading the crowded train. 

B2: Well, I suppose if you're absolutely sure my presence won't ruin your evening...  

B3: Nah, I could use the walk. Been sitting all day.

Intentions:  

  B1:  
  
    GM: None  
    
    Intention: Directly accepts and expresses gratitude. 
    Enthusiastic tone.  
               A speaker might use it to show appreciation and comfort.  

  B2:  
  
    GM: Quality  
    
    Intention: Uses sarcasm and self-deprecation. Adds humor while implying acceptance.  
               Useful for playful or noncommittal tone.  

  B3:  
  
    GM: Relation  
    
    Intention: Deflects the offer indirectly. Politely declines without confrontation.  
               Useful for maintaining autonomy while being polite.

\medskip

\textbf{6. JSON Output Requirements}  

- \texttt{"Context"}: The paraphrase of the given text, but do not shorten it.  

- \texttt{"Root"}: The initial statement (A’s opening statement).

- \texttt{"B1"}, \texttt{"B2"}, \texttt{"B3"}: The three replies to the root.  

- \texttt{"Intentions"}: Each reply must include \texttt{"GM"} and \texttt{"Intention"}.  

- Ensure responses are semantically similar but pragmatically distinct.  

- Do not output anything outside of the specified JSON format.

\medskip

\textbf{7. Final Instruction}  

When ready, check the conversation again to ensure all responses convey distinct pragmatic functions.  
If not, revise them and their associated intentions.  
\textbf{ONLY OUTPUT THE JSON FILE.} Do not include any explanations or additional commentary.

\end{tcolorbox}

\section{Prompt Templates for Response Generation}
\label{appendix:evaluation}

To obtain model's pragmatic analysis of each datapoint, we adopt the full assistant prompt format proposed by \citet{lin2024the}, including the instruction preamble and example completions. The only component we add is the final task-specific question shown below:

\begin{tcolorbox}[promptbox, title=Prompt Segment (Appended after \textsc{Urial} Prompt), colback=white, colframe=black!75, fonttitle=\bfseries]
\raggedright
Your task is to examine the following short conversation and assess: 

- What is the pragmatic intention behind `response\_1'? 

- Why or when might someone prefer `response\_1' over `response\_2' pragmatically? \\

Please answer in 1 paragraph. \\

Answer:
\end{tcolorbox}

\section{Hyperparameter Settings for Response Generation}
\label{appendix:hyperparameters}

To ensure comparability across models and avoid extraneous variance, we apply a unified set of decoding parameters for all generations, regardless of model architecture or size. The configuration is summarized below:

\begin{tcolorbox}[promptbox, title=Generation Settings (Applied to All Models), colback=white, colframe=black!75, fonttitle=\bfseries, left=2pt, right=2pt, top=4pt, bottom=4pt]
\begin{tabularx}{\columnwidth}{l l}
\texttt{max\_new\_tokens} & 256 \\
\texttt{do\_sample} & True \\
\texttt{top\_k} & 50 \\
\texttt{top\_p} & 1.0 \\
\texttt{temperature} & 0.5 \\
\texttt{repetition\_penalty} & 1.0 \\
\end{tabularx}
\end{tcolorbox}

This configuration was consistently applied to all evaluated models, including OLMo-2, OLMoE, LLaMA-3.1-Tülu-3, and Qwen-3 families.

\section{Prompt Templates for 10-Point Scoring}
\label{appendix:10-point-scoring}

In this setting, we evaluate the quality of intention explanations generated by models in different stages. To assess alignment with the intended pragmatic goal, we compare each model output to a human-annotated reference and ask GPT-4.1 to assign a score between 1 and 10, or ''Invalid'' if the response is incoherent. This helps filter out degenerate completions common in base models. Data points marked ''Invalid'' are excluded during score aggregation.

\begin{tcolorbox}[promptbox, title=Prompt Template for 10-Point Scoring, colback=white, colframe=black!75, fonttitle=\bfseries]
\raggedright
You are an expert evaluator of language model responses. Your task is to score how well a model's response captures the idea of a response to a given sentence.

\medskip
The tested model will see a conversation structure like this: \\
"context": ... \\
"initial turn": ... \\
"response\_1": ... \\
"response\_2": ...

And they are asked to evaluate: \\
What is the intention behind `response\_1'? Why or when might someone prefer `response\_1' over `response\_2'?

\medskip
We have the reference golden intention for you: ..., this is the core idea of the intention of response\_1.

Model's response: ...

You should understand that the tested models are base models, thus not deliberately instructed to format their responses and may output random strings and sentences. You just need to check the content of their responses instead of the format.

\medskip
Score the model's response on a scale of 1 to 10, or Invalid, where: \\
- 10: The model's response perfectly captures the intention described in the reference \\
- 1: The model's response poorly captures the intention described in the reference \\
- Invalid: The model's response is nonsense or invalid

Return ONLY a JSON object with the following format: \\
\{\{ \\
  "score": <number between 1 and 10 or Invalid>, \\
  "reason": "<brief explanation of your score, no more than 25 words>" \\
\}\}

Do not include any other text, just the JSON object.
\end{tcolorbox}

\section{Prompt Templates for Pairwise Comparison}
\label{appendix:pairwise-omparison}

We use the following prompt to evaluate two model explanations for a single datapoint. GPT-4.1 is asked to choose which model better captures the speaker’s intention and to classify the difference into one of three pragmatic categories. If either model's response is incoherent or if no clear winner can be determined, GPT-4.1 is instructed to return ''Invalid''.

\begin{tcolorbox}[promptbox, title=Prompt Template for Pairwise Comparison, colback=white, colframe=black!75, fonttitle=\bfseries]
\raggedright

You are an expert evaluator of language model analysis.
Your task is to compare two model analysis of the intention of response\_1, \\
which is a response to the "initial turn", and choose which one better aligns with \\
the golden label's intention.

\medskip
The conversation structure is: \\
"context": ... \\
"initial turn": ... \\
"response\_1": ... \\
"response\_2": ... \\
"golden\_label": ...

\medskip
The models are asked to evaluate: \\
What is the intention behind `response\_1'? \\
Why or when might someone prefer `response\_1' over `response\_2'?

\medskip
Model 1's response: ... \\
Model 2's response: ...

\medskip
Choose which response better captures the intention described in the golden label.

\medskip
After comparing the two responses, you also need to categorize how the better response \\
is better than the other.

\medskip
Here are the categories: \\

\textbf{Category \#1: Cognitive-Pragmatic Competence} \\
description: The better response goes beyond the literal meaning of the sentence and \\
identifies the speaker's underlying social goal, such as softening a refusal or \\
signaling indirect disagreement.

\medskip
\textbf{Category \#2: Pragmalinguistic Competence} \\
description: The better response identifies and explains rhetorical techniques—like humor, \\
irony, or self-deprecation—and clarifies how these strategies function to manage emotion \\
or social tension.

\medskip
\textbf{Category \#3: Sociopragmatic Competence} \\
description: The better response shows sensitivity to the social context, including roles, \\
relationships, or timing, and explains why the speaker's choice fits the situation appropriately.

\medskip
If you feel the better response does not fall into any of the categories, \\
you can choose "Invalid", and explain how it better than the other in the "reason" field.

\medskip
Return ONLY a JSON object with the following format: \\
\{\{ \\
\quad "choice": "1" or "2" or "Invalid", \\
\quad "reason": "\textless brief explanation in 20 words or less\textgreater", \\
\quad "category": "1" or "2" or "3" or "Invalid" \\
\}\}

\medskip
You should understand that the tested models might be base models, \\
thus not deliberately instructed to format their responses and may output \\
random strings and sentences. You need to purely focus on the content and \\
compare them instead of having preference on the format.

\medskip
Return "Invalid" if: \\
- Either response is empty, nonsense, unrelated to the question, or \\
  as long as you feel it is not answering the question \\
- You cannot make a clear choice between the two responses

\end{tcolorbox}

\section{Distributional statistics of 10-Point Scoring Evaluation}
\label{appendix:scoring-distribution}

We present the full distribution of all evaluated model variants (std, min, max, median) in \autoref{tab:appendix-scoring-distribution}. These results also suggest that DPO models usually show less fluctuation in interpretation. 

\begin{table}[h]
\centering
\small
\setlength{\tabcolsep}{4pt}
\renewcommand{\arraystretch}{0.9} 
\begin{tabularx}{\columnwidth}{lccccc}
\toprule
Model & Mean & Std & Min & Max & Median \\
\midrule
Llama-3-8B-Base & 7.01 & 2.03 & 1 & 10 & 8 \\
Llama-3-8B-SFT & 8.10 & 1.89 & 1 & 10 & 9 \\
Llama-3-8B-DPO & 8.48 & 1.68 & 1 & 10 & 9 \\
Llama-3-70B-Base & 8.26 & 1.65 & 1 & 10 & 9 \\
Llama-3-70B-SFT & 8.78 & 1.47 & 2 & 10 & 9 \\
Llama-3-70B-DPO & 9.28 & 1.12 & 2 & 10 & 10 \\
\midrule
OLMo-2-7B-Base & 6.13 & 2.19 & 1 & 10 & 6 \\
OLMo-2-7B-SFT & 7.66 & 1.99 & 1 & 10 & 8 \\
OLMo-2-7B-DPO & 8.52 & 1.68 & 1 & 10 & 9 \\
OLMo-2-13B-Base & 7.15 & 1.96 & 1 & 10 & 8 \\
OLMo-2-13B-SFT & 8.05 & 1.78 & 1 & 10 & 9 \\
OLMo-2-13B-DPO & 8.59 & 1.59 & 1 & 10 & 9 \\
OLMo-2-32B-Base & 8.04 & 1.70 & 2 & 10 & 9 \\
OLMo-2-32B-SFT & 8.47 & 1.58 & 1 & 10 & 9 \\
OLMo-2-32B-DPO & 8.90 & 1.35 & 1 & 10 & 9 \\
\midrule
OLMoE-1B-7B-Base & 3.45 & 2.18 & 1 & 10 & 3 \\
OLMoE-1B-7B-SFT & 5.29 & 2.51 & 1 & 10 & 6 \\
OLMoE-1B-7B-DPO & 5.96 & 2.59 & 1 & 10 & 7 \\
\midrule
Qwen3-0.6B & 4.46 & 2.38 & 1 & 10 & 4 \\
Qwen3-1.7B & 6.48 & 2.40 & 1 & 10 & 7 \\
Qwen3-4B & 7.66 & 2.11 & 1 & 10 & 8 \\
Qwen3-8B & 7.54 & 2.40 & 1 & 10 & 8 \\
\bottomrule
\end{tabularx}
\caption{Distributional statistics of 10-Point Scoring Evaluation.}
\label{tab:appendix-scoring-distribution}
\end{table}

\section{Examples from 10-Point Scoring Evaluation}
\label{appendix:scoring-results}

To better illustrate how we assess explanation quality in our 10-point scoring evaluation, we present representative examples of each score range (high, moderate, low) from the evaluation process. Each example includes the conversation context, potential responses, the human-annotated reference explanation (\texttt{response\_1\_intent}), the model-generated explanation, and the resulting evaluation score and rationale.

\begin{tcolorbox}[promptbox, title=Example of 10-Point Scoring Evaluation A: LLaMA-3.1-70B Base, colback=white, colframe=black!75, fonttitle=\bfseries]

\texttt{context:}
John is a teacher at an elementary school. When talking with the principal about a new student, who did poorly on her entrance examination, John said, "This one is really sharp."

\medskip
\texttt{initial\_turn:}  
This one is really sharp.

\medskip
\texttt{response\_1:}  
I think she will fit right in with some guidance.

\medskip
\texttt{response\_2:}  
She's just like any other student who needs a bit of encouragement.

\medskip
\texttt{response\_1\_intent:}  
This response does not flout any maxim. It aims to express optimism about the student’s ability to succeed with proper support and highlights that she has potential. A speaker might prefer this reply over response\_2 because it directly affirms the student's strengths and offers a solution, presenting the student in a positive and proactive light. This approach helps reinforce a sense of capability and possibility.

\medskip
\texttt{model\_response:}  
The pragmatic intention behind response\_1 is to express optimism and confidence in the new student's ability to succeed with some guidance. This statement implies that the student has potential and can thrive in the school environment with the right support.

Someone might prefer response\_1 over response\_2 pragmatically in situations where they want to emphasize the student's potential and the importance of providing guidance to help them succeed. response\_1 focuses on the student's strengths and the teacher's belief in their ability to fit in and do well, while response\_2 takes a more neutral stance and simply acknowledges that the student needs encouragement like any other student.

In summary, choosing response\_1 over response\_2 pragmatically can help communicate a more positive and supportive attitude towards the new student, highlighting their potential and the teacher's confidence in their ability to succeed with the right guidance.

\medskip
\texttt{scoring:}  

Score: 10 (high)

Reason: Accurately explains the optimistic intention, focus on student potential, and why one might prefer this over a more neutral response.

\end{tcolorbox}

\begin{tcolorbox}[promptbox, title=Example of 10-Point Scoring Evaluation B: Olmo2-7B Base, colback=white, colframe=black!75, fonttitle=\bfseries]

\texttt{context:}  
Callie promised to bring a salad to Jason's dinner party. She arrived two hours late. Jason said, "I'm so glad that you were able to fit us into your busy schedule."

\medskip
\texttt{initial\_turn:}  
I'm so glad that you were able to fit us into your busy schedule.

\medskip
\texttt{response\_1:}  
Well, it's a good thing I'm fashionably late, right?

\medskip
\texttt{response\_2:}  
I know, right? Had to fight through traffic, but I'm here now.

\medskip
\texttt{response\_1\_intent:}  
This response flouts the maxim of Quality by playfully embracing the notion of being "fashionably late" instead of seriously addressing the tardiness. It aims to use humor and irony to diffuse any awkwardness and avoid direct confrontation. A speaker might prefer this reply over response\_2 because it turns the situation into a joke, attempting to lighten the mood rather than provide an explanation. This approach could help smooth over tension if the relationship is friendly enough for such banter.

\medskip
\texttt{model\_response:}  
In this context, the pragmatic intention behind the response\_1 is to express gratitude and appreciation for the other person's effort in accommodating the delay. This is because the speaker, Jason, is acknowledging that Callie was able to attend the dinner party despite being late. The speaker may prefer response\_1 over response\_2 pragmatically because it is more lighthearted and humorous, which can help to ease the tension or awkwardness that often arises from being late. Additionally, response\_1 does not explicitly acknowledge the reason for the delay, which may be considered more polite in certain contexts. 
On the other hand, response\_2 explicitly states the reason for the delay, which can be seen as more direct and assertive. This can be more appropriate in certain situations, such as when the speaker wants to emphasize the importance of punctuality or when they are trying to avoid any misunderstandings about the reason for the delay. Overall, the pragmatic intention behind response\_1 is to express gratitude and appreciation for the other person's effort, and it may be preferred in certain situations due to its lighthearted and humorous tone.

\medskip
\texttt{scoring:}  

Score: 5 (moderate)

Reason: Captures the humor and lightheartedness, but incorrectly says the intention is gratitude/appreciation, missing the playful, ironic deflection central to the reference.

\end{tcolorbox}

\begin{tcolorbox}[promptbox, title=Example of 10-Point Scoring Evaluation C: Olmo2-7B Base, colback=white, colframe=black!75, fonttitle=\bfseries]

\texttt{context:}
X wants to know what types of books Y prefers to read.

\medskip
\texttt{initial\_turn:}  
Do you enjoy mysteries?

\medskip
\texttt{response\_1:}  
I guess you could say I like getting my mind tangled up in a bit of a puzzle now and then.

\medskip
\texttt{response\_2:}  
Oh yes, I love the suspense and twists they offer!

\medskip
\texttt{response\_1\_intent:}  
This response flouts the maxim of Quality. It aims to convey enjoyment of mysteries indirectly through playful understatement, suggesting interest by framing it as enjoying “getting my mind tangled up in a puzzle.” A speaker might prefer this reply over response\_2 because it uses humor and a light tone to keep the conversation casual or to downplay enthusiasm in a friendly way. This adds a touch of personality and playfulness to the exchange.

\medskip
\texttt{model\_response:}  
In this dialogue, X asks Y whether they enjoy reading mysteries. Y's response, "I guess you could say I like getting my mind tangled up in a bit of a puzzle now and then," indicates that they enjoy reading mystery novels. This response is pragmatic because it provides information about Y's preferences and interests, which is relevant to the context of the conversation. Y's response, "Oh yes, I love the suspense and twists they offer!" is also pragmatic because it provides additional information about their preferences. However, this response is more specific and direct, indicating that Y enjoys the suspense and twists that mystery novels offer. 

Therefore, Y might prefer "Oh yes, I love the suspense and twists they offer!" over "I guess you could say I like getting my mind tangled up in a bit of a puzzle now and then" pragmatically when they want to be more clear and direct about their preferences.

\medskip
\texttt{scoring:}  

Score: 3 (low)

Reason: The response misses the playful, understated intention and focus on tone; it only notes indirectness and informativeness, not the humor or personality.

\end{tcolorbox}

\section{Examples from Pairwise Comparison Evaluation}
\label{appendix:comparison-results}

To further illustrate how different models perform in pragmatic reasoning, we present selected examples from our pairwise comparison evaluation. In this setting, two model-generated responses are compared against a human-annotated pragmatic interpretation (\texttt{response\_1\_intention}) to determine which aligns better with the intended meaning. For the winning explanation, we further categorize it into one of three pragmatic dimensions as discussed in Section~\ref{sec:eval-metrics}.

We provide one illustrative example per category to showcase the types of reasoning improvements observed in our evaluations.

\begin{tcolorbox}[promptbox, title=Pairwise Comparison Example A: LLaMA-3.1-70B Base vs SFT, colback=white, colframe=black!75, fonttitle=\bfseries]

\texttt{context:}  
A boy from New York was being led through the swamps of Florida and is curious about whether alligators will attack if he carries a flashlight.

\medskip
\texttt{initial\_turn:}  
Is it true that an alligator won't attack you if you carry a flashlight?

\medskip
\texttt{response\_1:}  
Well, let's just say alligators have their own mind and aren't that predictable.

\medskip
\texttt{response\_2:}  
That's actually just a myth, but keeping one handy is a good idea for other reasons.

\medskip
\texttt{response\_1\_intention:}  
This response flouts the maxim of Relation. It aims to answer indirectly, emphasizing the unpredictability of alligators rather than directly addressing the myth. A speaker might prefer this reply over response\_2 because it introduces a sense of caution and intrigue, perhaps leaving the listener more alert or curious. The indirectness could serve to prompt further questions or reinforce respect for wildlife unpredictability.

\medskip
\texttt{model\_1\_response:}  
The pragmatic intention behind `response\_1' is to express uncertainty, skepticism, or a lack of definitive knowledge about the behavior of alligators. By saying "alligators have their own mind and aren't that predictable," the speaker is suggesting that it's difficult to make general statements about how alligators will react in specific situations. This may be a more appropriate response if the speaker is unsure of the veracity of the claim or wants to avoid making a definitive statement about something they don't know for certain.  

Someone might prefer `response\_1' over `response\_2' pragmatically in situations where they want to express uncertainty or skepticism, or when they don't want to make a definitive statement about something they aren't sure of. This could be particularly relevant in the context of the conversation, where the speaker is being asked about a potentially dangerous situation and may want to avoid giving misleading information.

\medskip
\texttt{model\_2\_response:}  
The pragmatic intention behind `response\_1' is to express uncertainty and caution about the predictability of alligators' behavior. The speaker may prefer `response\_1' over `response\_2' when they want to emphasize the unpredictable nature of alligators and the importance of being cautious in general, rather than focusing on the specific myth about flashlights. This approach may be more appropriate when the speaker wants to encourage the listener to be vigilant and aware of potential dangers in the swamps, regardless of the presence of a flashlight.

\medskip
\texttt{comparison\_result:}  

Choice: 2  

Reason: Model 2 highlights indirectness and caution, matching the golden label's focus on unpredictability and vigilance.

Category: 1

\end{tcolorbox}

\begin{tcolorbox}[promptbox, title=Pairwise Comparison Example B: LLaMA-3.1-8B Base vs SFT, colback=white, colframe=black!75, fonttitle=\bfseries]

\texttt{context:}  
X wants to know about Y's food preferences by asking if Y enjoys Chinese cuisine.

\medskip
\texttt{initial\_turn:}  
Are you into Chinese cuisine?

\medskip
\texttt{response\_1:}  
Well, I guess I've survived a couple of meals.

\medskip
\texttt{response\_2:}  
It's okay, but I'm more of a Mediterranean fan.

\medskip
\texttt{response\_1\_intention:}  
This response flouts the maxim of Quality by using sarcasm to imply only a minimal or reluctant enjoyment of Chinese food. It aims to add humor and ambiguity, downplaying direct expression of preference. A speaker might prefer this reply over response\_2 to keep the conversation playful and avoid direct criticism, especially if they want to maintain a light tone.

\medskip
\texttt{model\_1\_response:}  
The pragmatic intention behind `response\_1' is to provide a noncommittal answer to the question about liking Chinese cuisine. It is a way to avoid giving a direct answer, which might be interpreted as a personal preference. `response\_1' is preferred over `response\_2' pragmatically because it offers a more neutral and noncommittal response. By saying "I guess I've survived a couple of meals," the speaker is implying that they have had some experience with Chinese cuisine, but they are not necessarily a fan of it. This response allows the speaker to save face and avoid appearing too negative about Chinese cuisine, while still conveying that they may not be a big fan. Additionally, it leaves room for the conversation to continue without being closed off by a definitive answer.

\medskip
\texttt{model\_2\_response:}  
The pragmatic intention behind `response\_1' is to express a lack of strong preference or enthusiasm for Chinese cuisine, possibly indicating a neutral or indifferent attitude towards it. Someone might prefer `response\_1' over `response\_2' pragmatically in a situation where they do not want to completely rule out the possibility of enjoying Chinese food, but also do not want to commit to a strong positive stance. `response\_1' could be considered a middle ground or a way to provide a non-committal response, which might be a preferred strategy in polite conversation or when the speaker does not want to appear overly critical or dismissive of the food. This response can be useful in maintaining a neutral tone and avoiding potential awkwardness or conflict in the conversation.

\medskip
\texttt{comparison\_result:}  

Choice: 1  

Reason: Model 1 recognizes the use of humor and indirectness, aligning with the golden label's focus on sarcasm and playfulness.

Category: 2

\end{tcolorbox}

\begin{tcolorbox}[promptbox, title=Pairwise Comparison Example C: LLaMA-3.1-7B SFT vs DPO, colback=white, colframe=black!75, fonttitle=\bfseries]

\texttt{context:}  
On his first flight while learning to skydive, a man nervously anticipated his first jump, so he said to his instructor, "I packed the parachute myself, but I'm sure it won't open."

\medskip
\texttt{initial\_turn:}  
I packed the parachute myself, but I'm sure it won't open.

\medskip
\texttt{response\_1:}  
Don't worry, we've double-checked it. You're in safe hands.

\medskip
\texttt{response\_2:}  
You'll be just fine. And if not, it's been nice knowing you!

\medskip
\texttt{response\_1\_intention:}  
This response does not flout any maxim. It aims to provide direct reassurance to the nervous jumper by emphasizing that the equipment has been double-checked and he is safe. A speaker might prefer this reply over response\_2 because it maintains a calming and professional tone, offering clear comfort and building trust in a potentially stressful situation.

\medskip
\texttt{model\_1\_response:}  
The pragmatic intention behind "Don't worry, we've double-checked it. You're in safe hands." is to reassure the speaker's partner that he is in good care and that the parachute has been thoroughly checked, which would be preferred over response\_2 pragmatically because it provides a sense of security and confidence in the safety measures that are in place, which is important in a potentially dangerous situation like skydiving. This sentence would be more comforting to the partner and would also serve to alleviate any concerns the partner may have about the safety of the parachute.

\medskip
\texttt{model\_2\_response:}  
The pragmatic intention behind the sentence "Don't worry, we've double-checked it. You're in safe hands." (`response\_1') is to provide reassurance and comfort to the nervous skydiver. This sentence is chosen to pragmatically offer support and confidence, alleviating the skydiver's anxiety about the parachute. It's likely preferred because it communicates a sense of professionalism and thoroughness in safety checks, thereby reassuring the skydiver that all precautions have been taken. This sentence is appropriate when someone needs to be comforted and needs to feel secure in a potentially risky situation. In contrast, "You'll be just fine. And if not, it's been nice knowing you!" (`response\_2') is a more lighthearted and possibly flippant response that might not provide the necessary comfort or professionalism that the skydiver is seeking in this context. While `response\_2' might be used in a more joking or casual setting, it lacks the assurance and emphasis on safety that the skydiver needs to feel secure about jumping out of a plane.

\medskip
\texttt{comparison\_result:}  

Choice: 2  

Reason: Model 2 explains both reassurance and professionalism, matching the golden label's emphasis on tone and trust.

Category: 3

\end{tcolorbox}

\section{Where Do SFT and DPO Help? A Case Study}
\label{appendix:case-study}

Here we present a full example from our dataset along with responses generated by the OLMo-2 7B model at three different training stages: Base, SFT, and DPO, accompanied by a detailed analysis.

We begin by comparing the Base and SFT versions. The Base model's explanation correctly identifies the intention of the sentence—namely, the "willingness to help"—but does not explore why the speaker chose this particular phrasing. The SFT version moves beyond literal intent by uncovering the speaker’s underlying communicative strategy: "conveys a sense of eagerness to assist and provides a quick solution to the problem." This recognition is critical—not simply that help is offered, but that the speaker chooses a form that signals initiative and decisiveness. The explanation frames this as a deliberate effort to convey responsiveness and reliability: "when the speaker wants to show their willingness to assist without waiting for confirmation." This shift from what is said to how and why it is said marks an important step toward cognitive-pragmatic reasoning, where language is understood as serving a strategic communicative function. In addition, the Base explanation includes a subtle misinterpretation of response\_2’s intention, claiming it is "not necessarily offering to help move it", which is clearly inaccurate, as response\_2 explicitly states "I’ll move it now". What differs is not the willingness to act, but the tone and framing: the speaker begins with a defensive justification (“Didn’t realize it was in your way”) before committing to the action. The SFT version avoids this error by simply skipping the interpretation of response\_2 and correctly characterizing response\_1 as more proactive and time-aware.

DPO’s response builds on the strengths of the SFT output by showing a more refined grasp of sociopragmatic competence, particularly in recognizing speaker–listener expectations and the timing of interpersonal actions. While the SFT model notes contextual factors like “when time is of the essence” and “without waiting for confirmation,” its explanation remains relatively speaker-centered and focused on efficiency. In contrast, the DPO explanation shows greater awareness of the interpersonal dimension. It describes response\_1 as “a direct and supportive response,” highlighting not just its immediacy, but how it reassures the listener and supports the smooth progression of the task. By noting that it “makes the speaker appear more proactive and helpful,” DPO links language to impression management and relational goals—key components of sociopragmatic reasoning. DPO also draws a sharper contrast between the candidates. It observes that response\_2 “first acknowledges the oversight and then offers to fix it,” introducing a slight delay that may signal inattentiveness or misalignment with the listener’s immediate needs. Thus, DPO shows that response\_1 is not just quicker, but also more socially attuned. Notably, its mention of contexts “where the speaker aims to establish themselves as a reliable and considerate helper” captures considerations of listener expectations, relationship dynamics, and situational urgency—elements entirely absent in the Base version and only lightly touched on in the SFT version.

Below is the full example along with responses:

\begin{tcolorbox}[promptbox, title=Responses for the Case Study: OLMo-2-1124-7B Base/SFT/DPO, colback=white, colframe=black!75, fonttitle=\bfseries]

\texttt{context:}  
A house painter is carrying a heavy tray of paints. A small ladder is blocking the sidewalk and the painter cannot get around it. He calls out to his brother: Can you move the ladder?

\medskip
\texttt{initial\_turn:}  
Can you move the ladder?

\medskip
\texttt{response\_1:} 
Sure thing, I'll get it out of your way right now.

\medskip
\texttt{response\_2:} 
Didn't realize it was in your way. I'll move it now.

\medskip
\texttt{response\_1\_intention:} 
This response does not flout any maxim. It aims to provide direct and enthusiastic agreement to the request, assuring the speaker that the ladder will be moved immediately. A speaker might prefer this reply over response\_2 because it conveys readiness and eagerness to help without any additional context, making it more efficient and supportive in an urgent situation.

\medskip
\texttt{response\_Base:} 
The pragmatic intention behind response\_1 "Sure thing, I'll get it out of your way right now" is to convey the speaker's willingness to assist in moving the ladder out of the painter's way. This sentence is preferred over response\_2 "Didn't realize it was in your way. I'll move it now" because it is more direct and explicit in indicating the speaker's willingness to help. In contrast, response\_2 "Didn't realize it was in your way. I'll move it now" could be interpreted as the speaker acknowledging the inconvenience caused by the ladder, but not necessarily offering to help move it. The speaker may be indicating that they will move the ladder, but this is not explicitly stated, which could lead to confusion or misunderstanding. Therefore, response\_1 "Sure thing, I'll get it out of your way right now" is more appropriate in this context because it clearly communicates the speaker's intention to help.

\medskip
\texttt{response\_SFT:} 
The pragmatic intention behind response\_1 is to affirm the speaker's willingness to help by using an immediate and affirmative response. This sentence conveys a sense of eagerness to assist and provides a quick solution to the problem. If someone prefers response\_1 pragmatically, they might be looking for a prompt action to resolve the issue at hand. In this case, the speaker is acknowledging the need for the ladder to be moved and is offering to take care of it right away. This can be seen as more helpful and proactive, which may be preferred in situations where time is of the essence or when the speaker wants to show their willingness to assist without waiting for confirmation.

\medskip
\texttt{response\_DPO:} 
The pragmatic intention behind response\_1 ("Sure thing, I'll get it out of your way right now.") is to affirm the speaker's willingness and immediate action to help the listener, which is a direct and supportive response to the request. This sentence conveys a sense of eagerness and promptness in addressing the problem, potentially making the speaker appear more proactive and helpful. On the other hand, response\_2 ("Didn't realize it was in your way. I'll move it now.") first acknowledges the oversight and then offers to fix it, which, while still cooperative, does so with a slight delay in recognizing the issue. Pragmatically, someone might prefer response\_1 over response\_2 when they want to emphasize their immediate readiness and willingness to assist without delay, especially in situations where time is of the essence or when the speaker wishes to express their proactive nature. This choice might be preferred in contexts where the listener values prompt action and direct assurance of support, such as in emergencies or when the speaker aims to establish themselves as a reliable and considerate helper.
\end{tcolorbox}

\section{Pairwise Comparison Category Distributions}
\label{appendix:end-to-end-prag}

To supplement our main findings, we present the full set of pairwise comparison results across all evaluated model pairs. Each figure below visualizes the distribution of winning explanations across three pragmatic competence categories. 

\begin{figure}[h]
    \centering
    \includegraphics[width=0.8\linewidth]{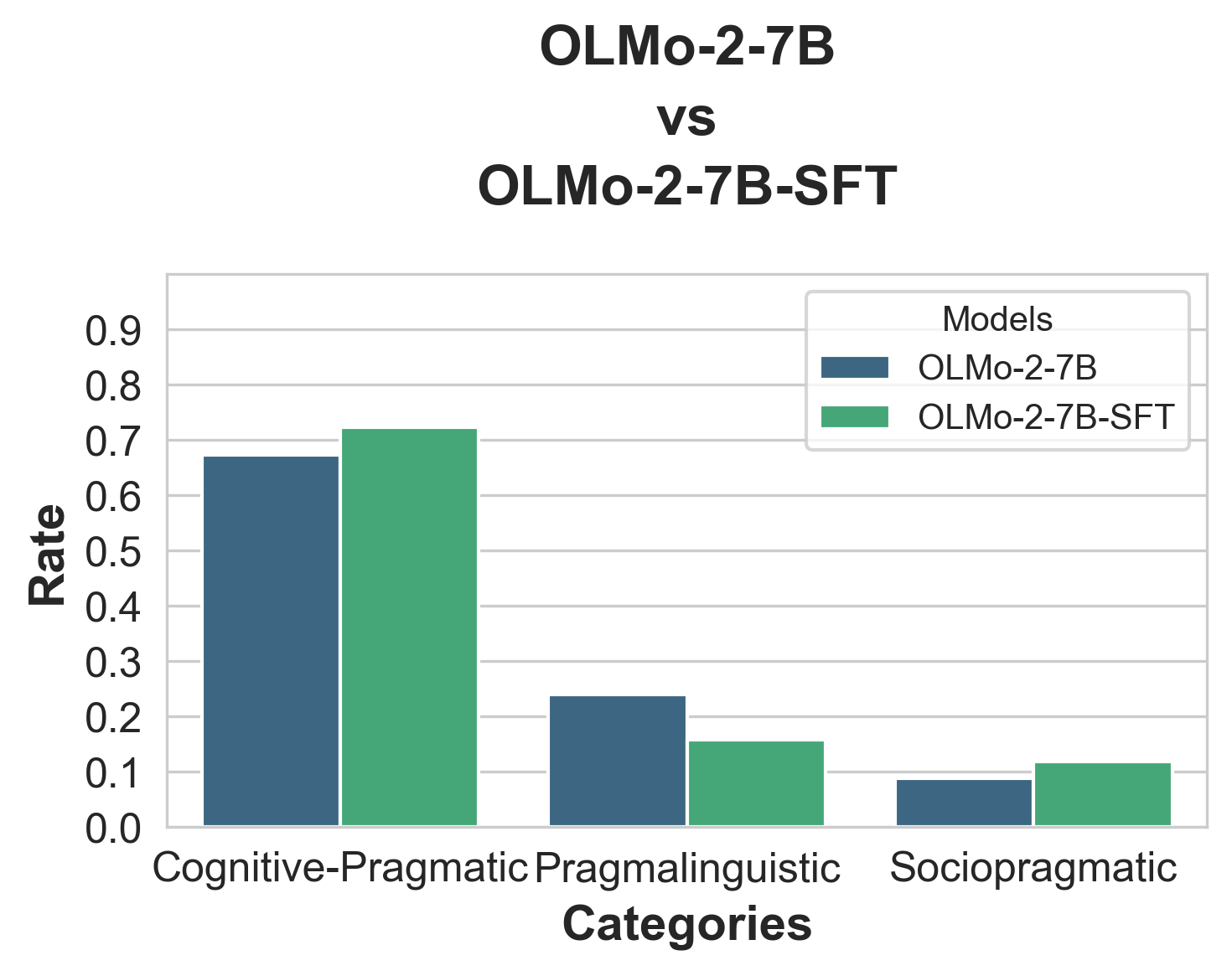}
    \caption{OLMo-2-7B Base vs SFT.}
    \label{fig:1}
\end{figure}
\begin{figure}[h]
    \centering
    \includegraphics[width=0.8\linewidth]{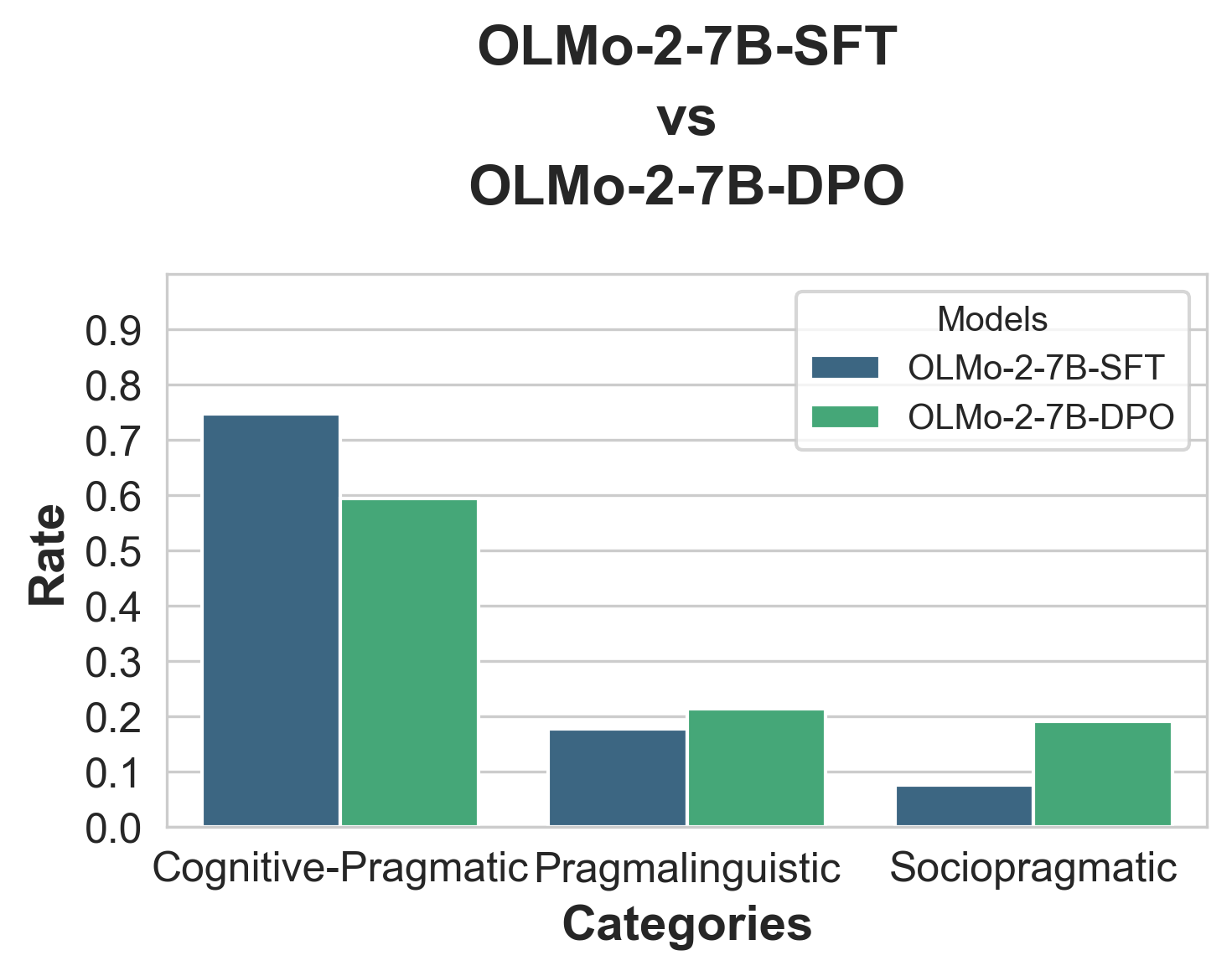}
    \caption{OLMo-2-7B SFT vs DPO.}
    \label{fig:2}
\end{figure}
\begin{figure}[h]
    \centering
    \includegraphics[width=0.8\linewidth]{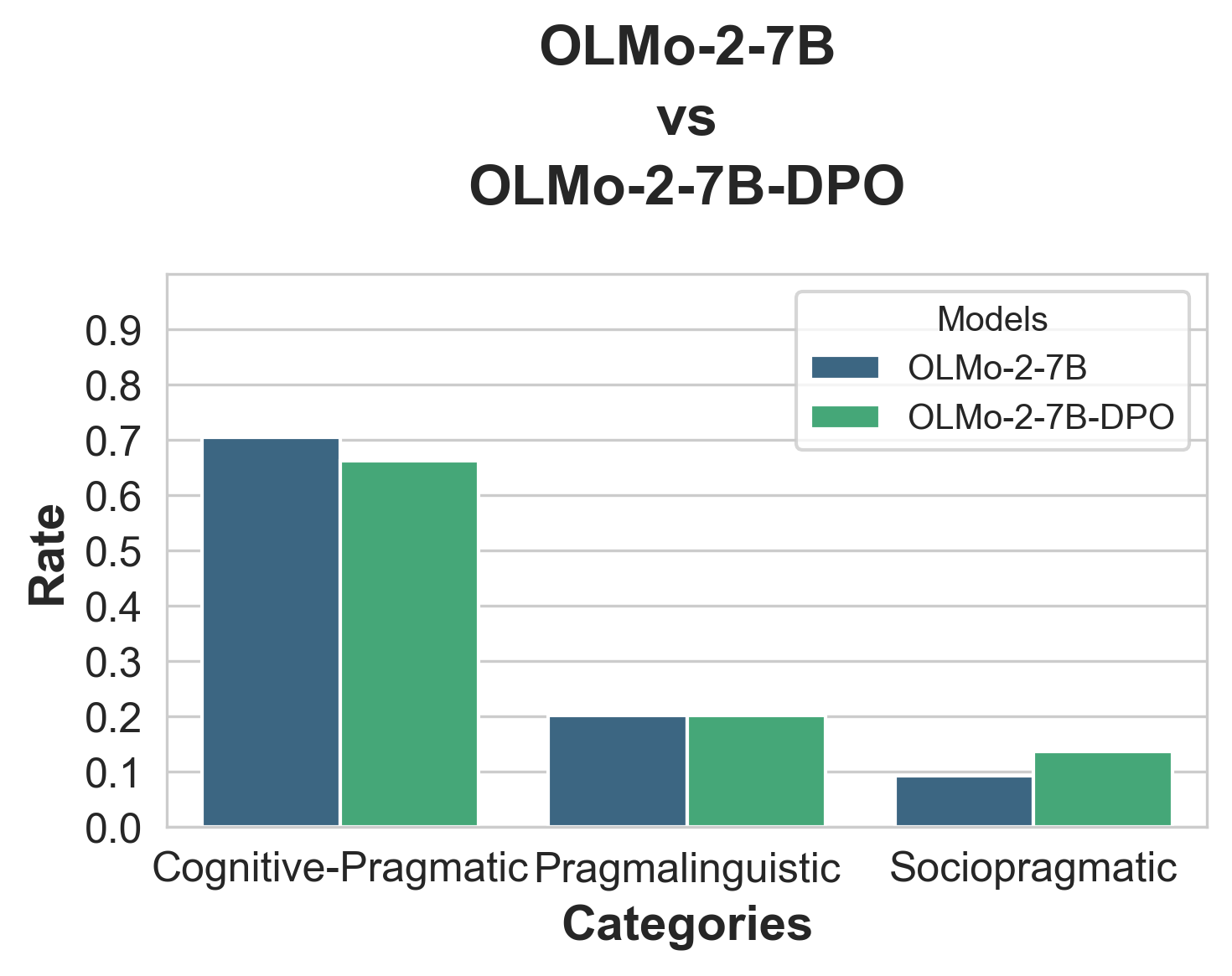}
    \caption{OLMo-2-7B Base vs DPO.}
    \label{fig:3}
\end{figure}
\begin{figure}[h]
    \centering
    \includegraphics[width=0.8\linewidth]{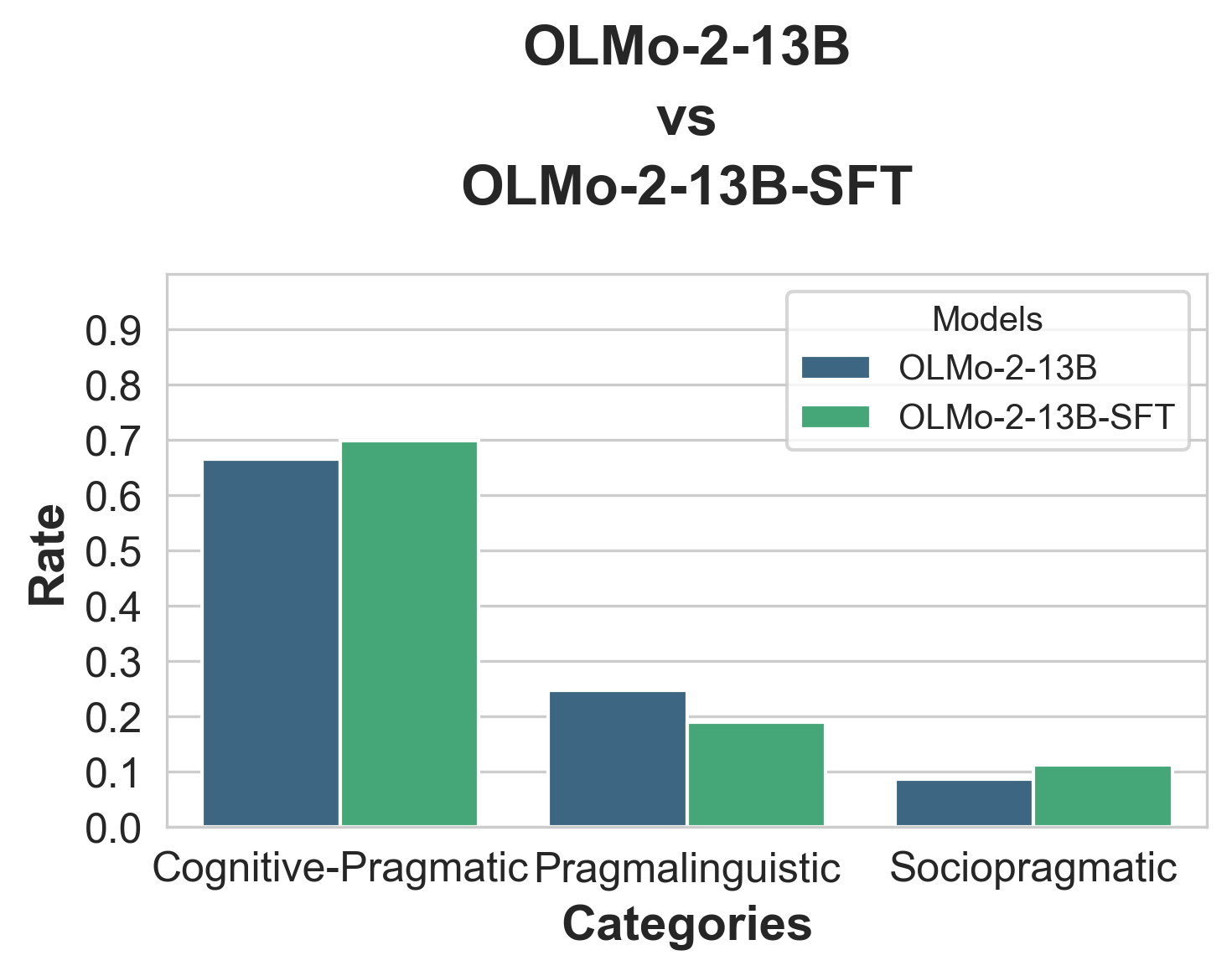}
    \caption{OLMo-2-13B Base vs SFT.}
    \label{fig:4}
\end{figure}
\begin{figure}[h]
    \centering
    \includegraphics[width=0.8\linewidth]{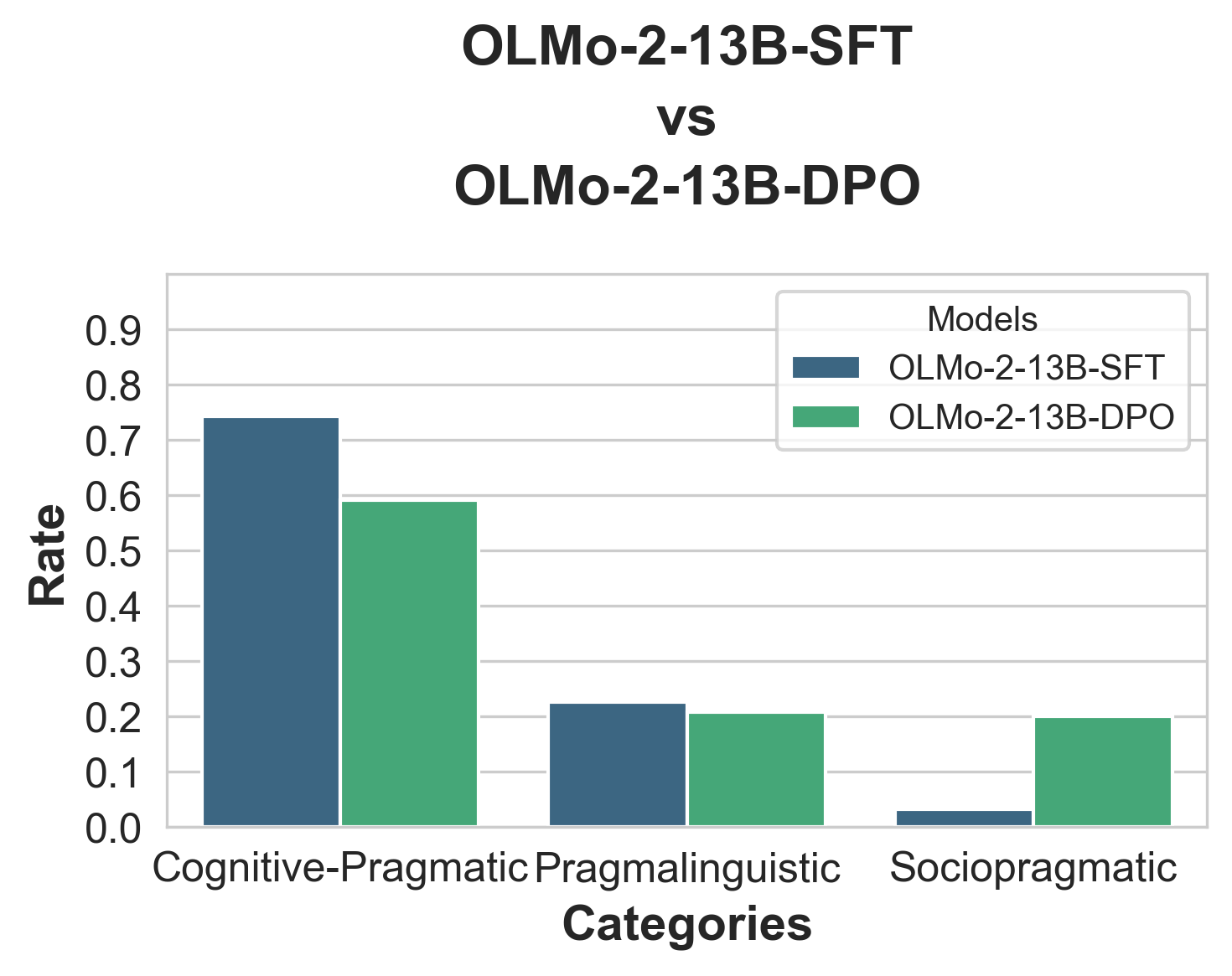}
    \caption{OLMo-2-13B SFT vs DPO.}
    \label{fig:5}
\end{figure}
\begin{figure}[h]
    \centering
    \includegraphics[width=0.8\linewidth]{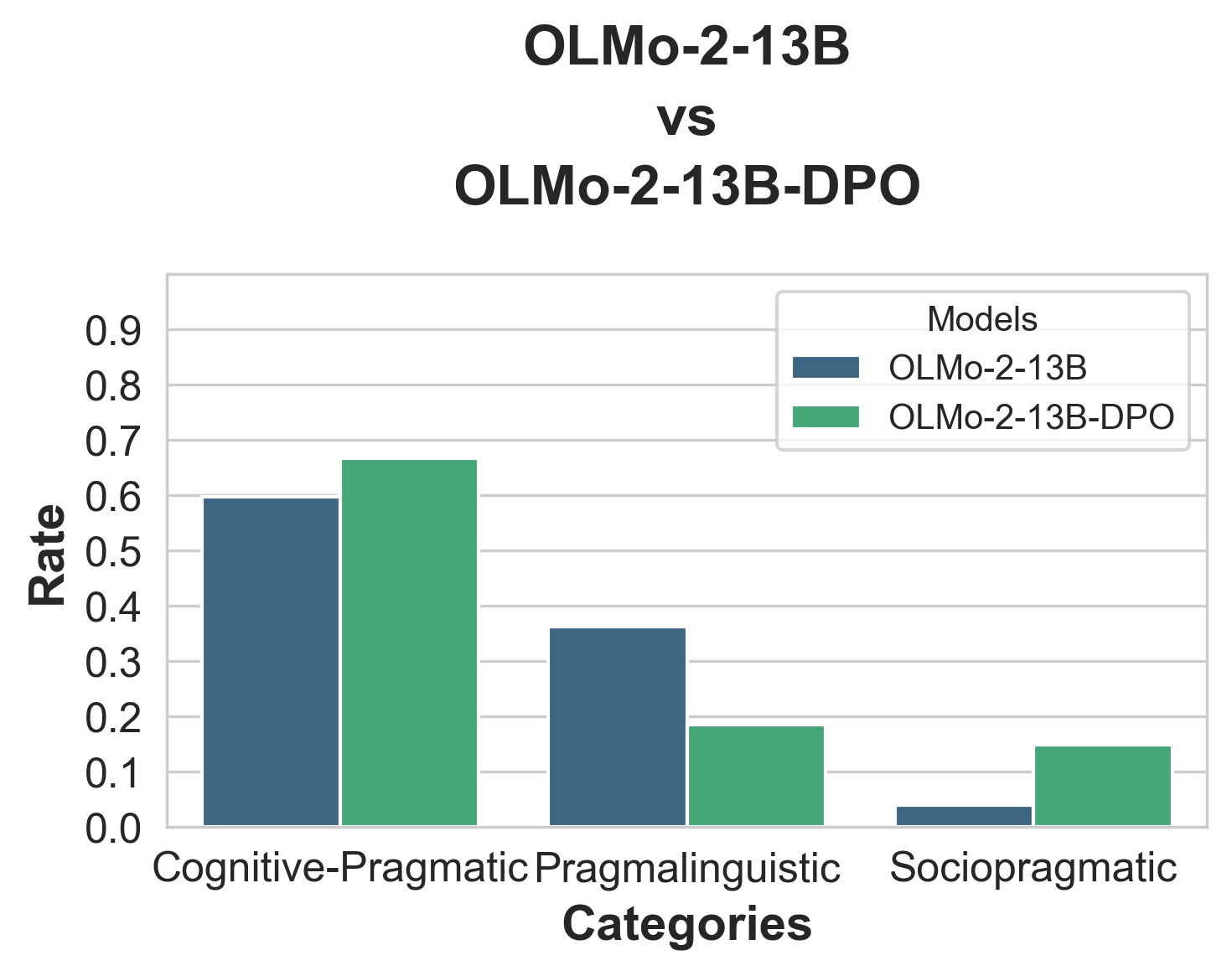}
    \caption{OLMo-2-13B Base vs DPO.}
    \label{fig:6}
\end{figure}
\begin{figure}[h]
    \centering
    \includegraphics[width=0.8\linewidth]{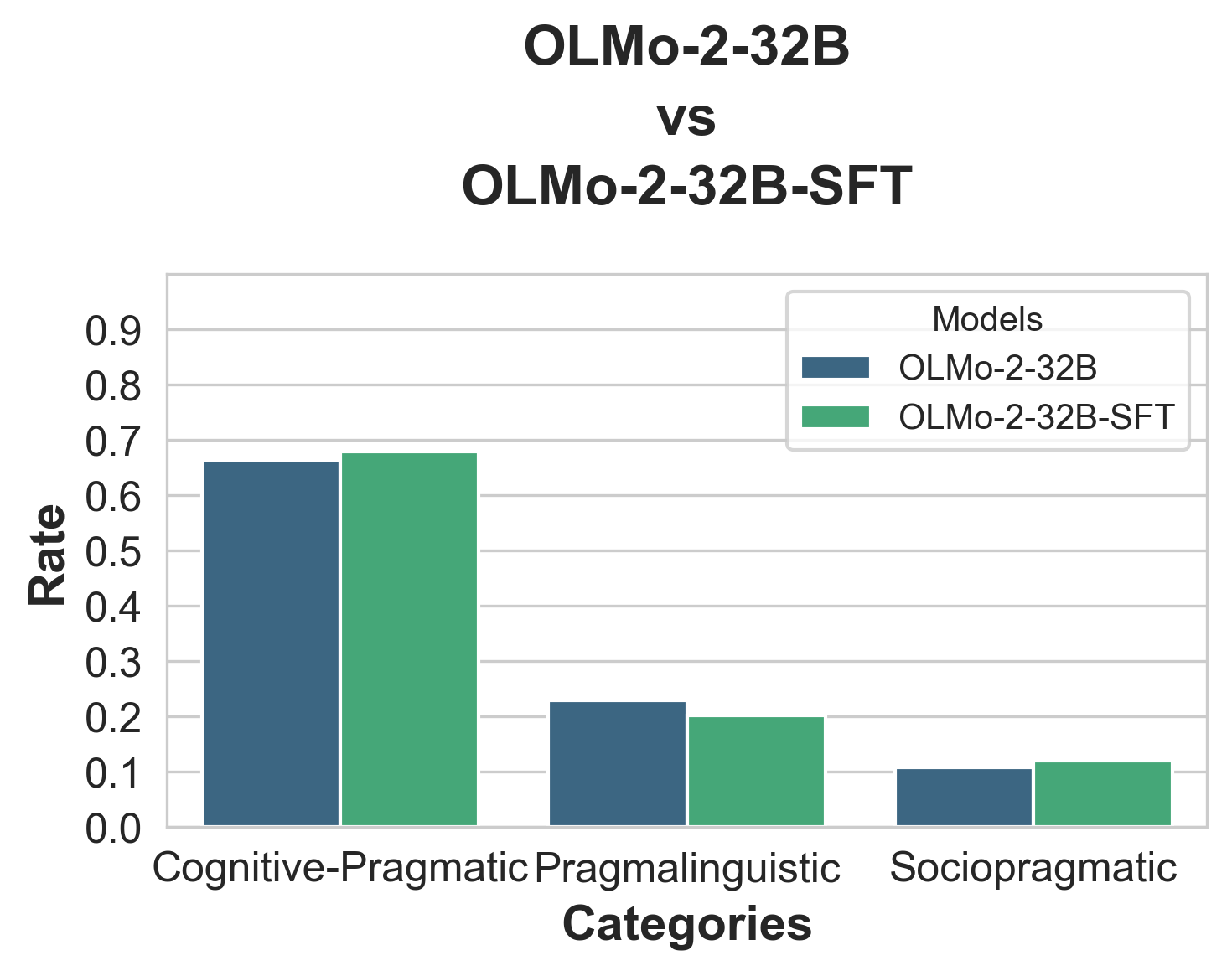}
    \caption{OLMo-2-32B Base vs SFT.}
    \label{fig:7}
\end{figure}
\begin{figure}[h]
    \centering
    \includegraphics[width=0.8\linewidth]{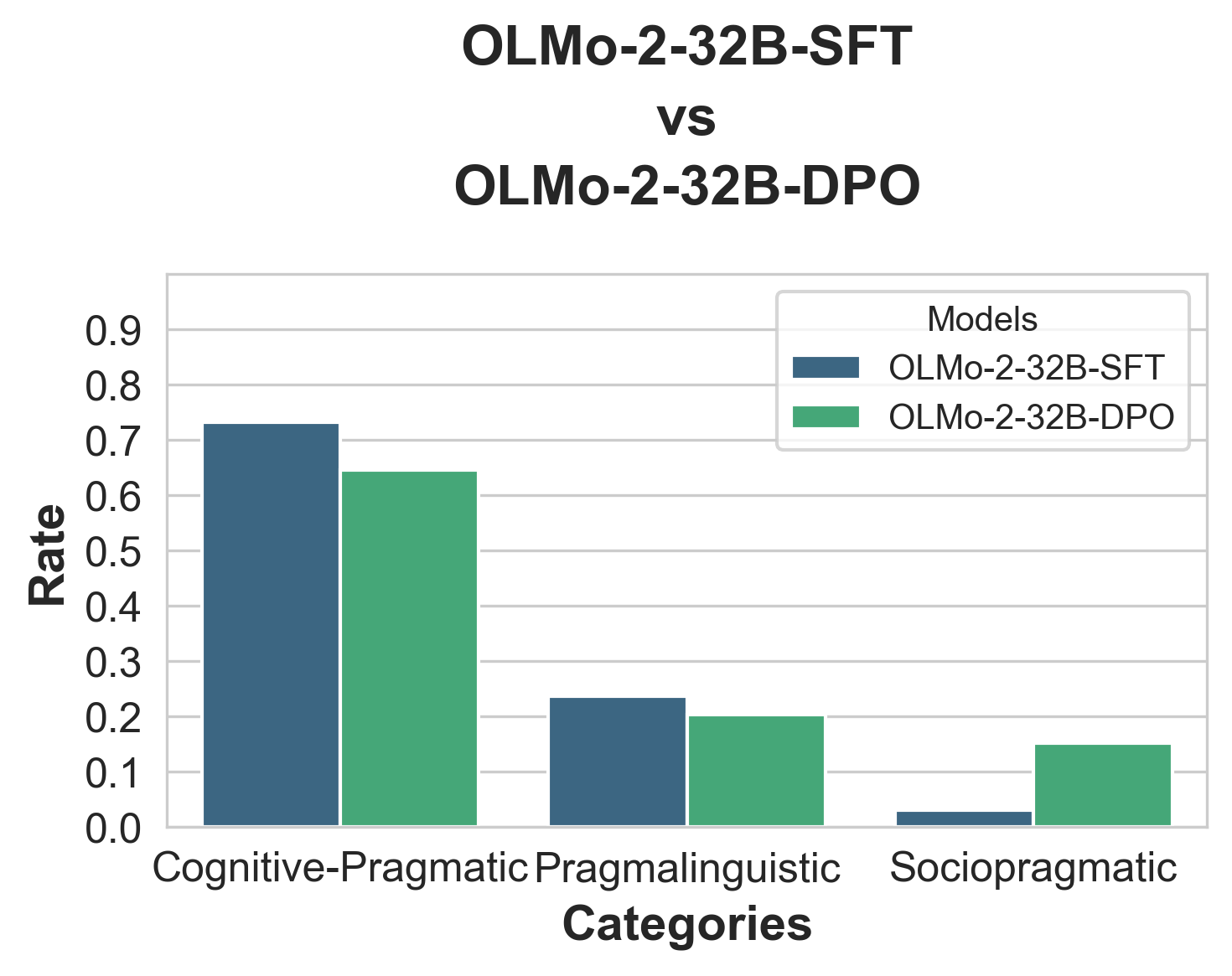}
    \caption{OLMo-2-32B SFT vs DPO.}
    \label{fig:8}
\end{figure}
\begin{figure}[h]
    \centering
    \includegraphics[width=0.8\linewidth]{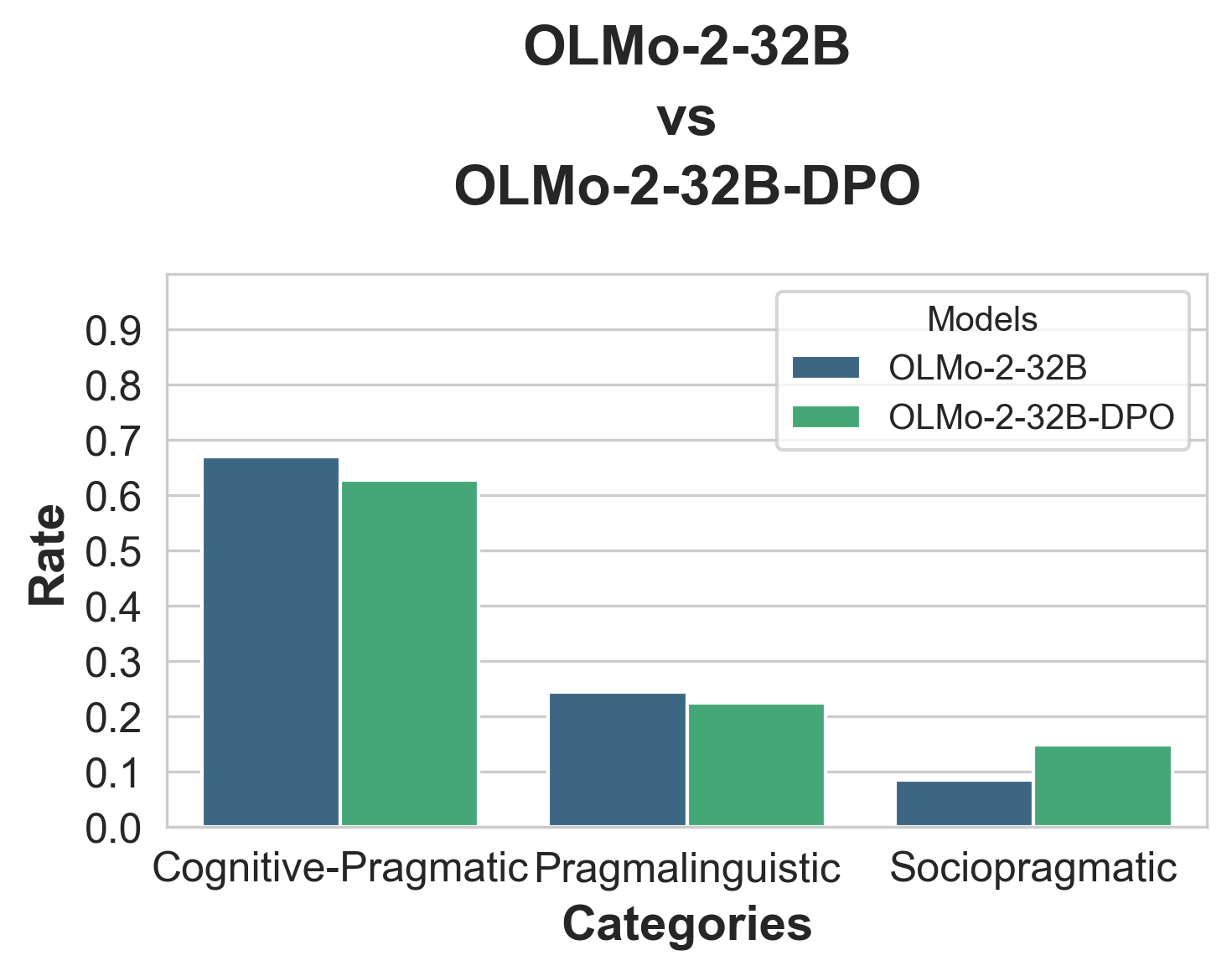}
    \caption{OLMo-2-32B Base vs DPO.}
    \label{fig:9}
\end{figure}
\begin{figure}[h]
    \centering
    \includegraphics[width=0.8\linewidth]{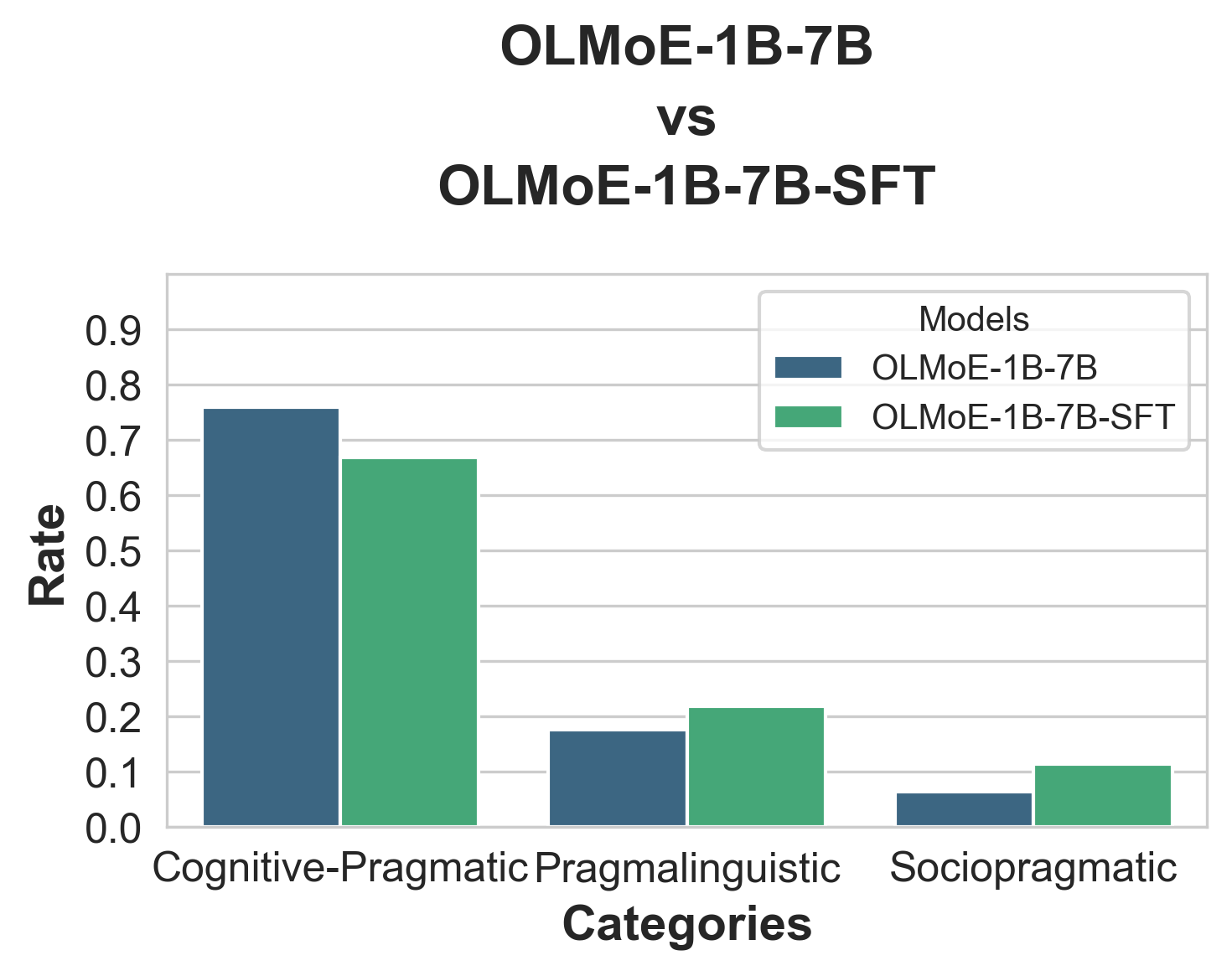}
    \caption{OLMoE-1B-7B Base vs SFT.}
    \label{fig:10}
\end{figure}
\begin{figure}[h]
    \centering
    \includegraphics[width=0.8\linewidth]{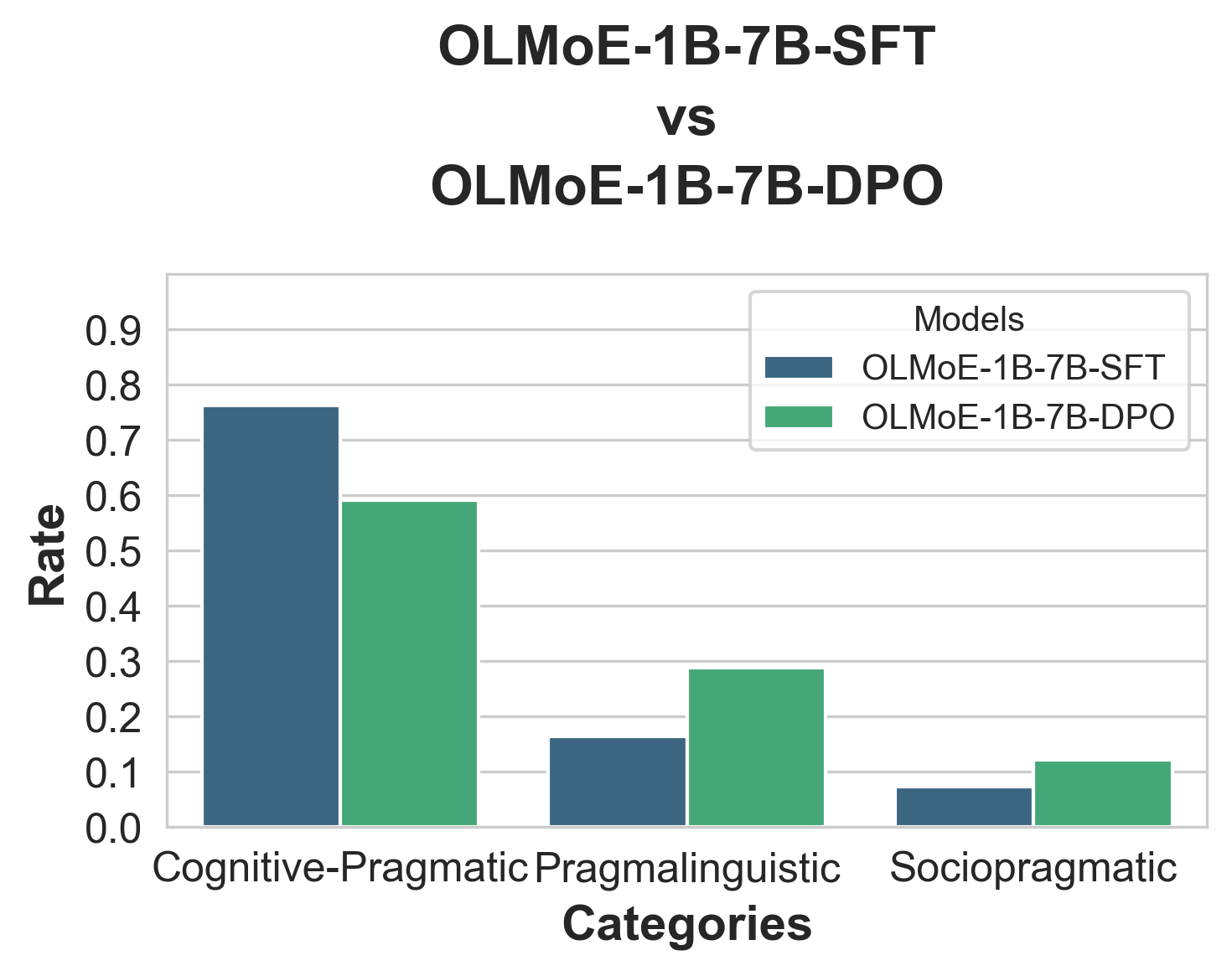}
    \caption{OLMoE-1B-7B SFT vs DPO.}
    \label{fig:11}
\end{figure}
\begin{figure}[h]
    \centering
    \includegraphics[width=0.8\linewidth]{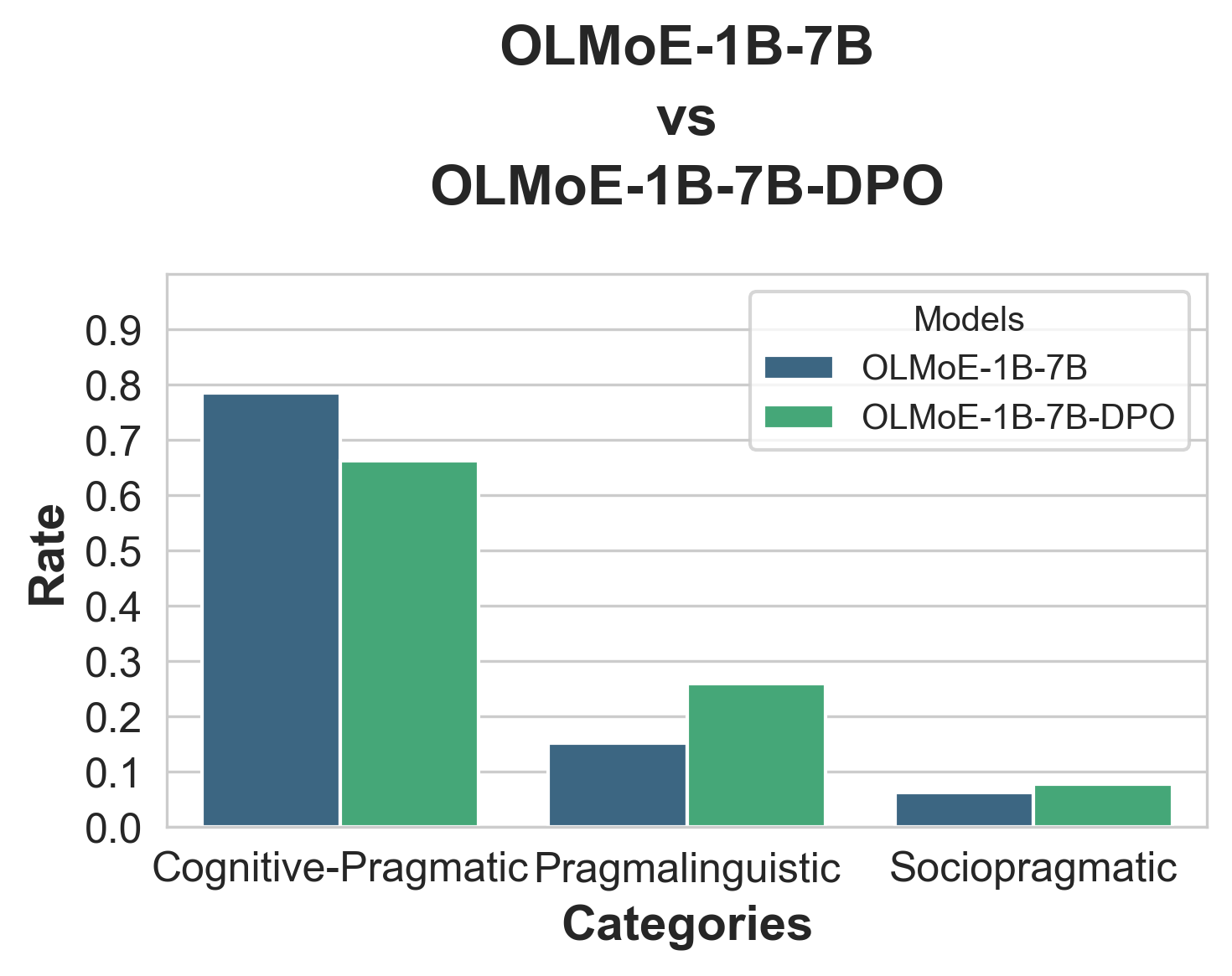}
    \caption{OLMoE-1B-7B Base vs DPO.}
    \label{fig:12}
\end{figure}
\begin{figure}[h]
    \centering
    \includegraphics[width=0.8\linewidth]{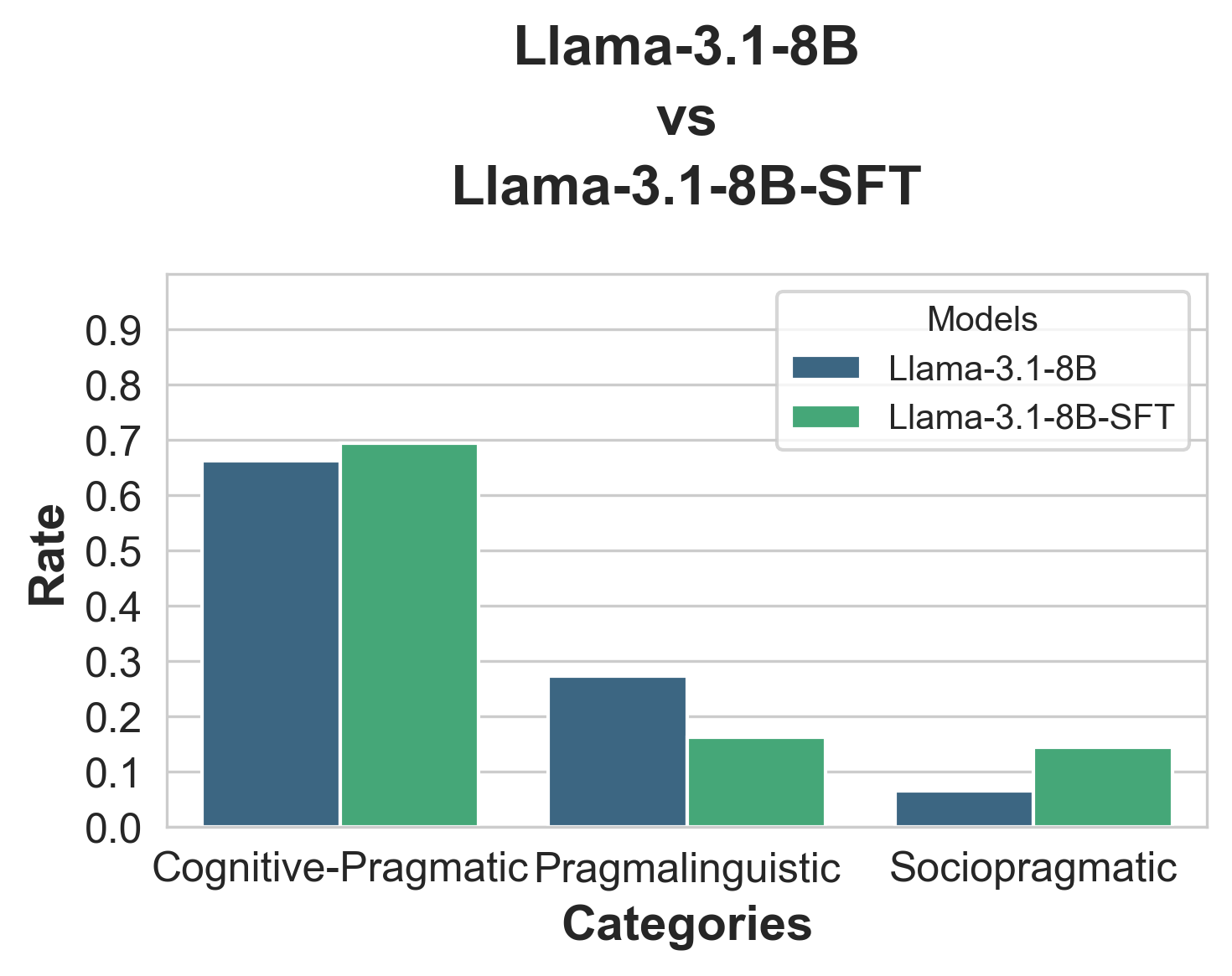}
    \caption{LLaMA-3.1-8B Base vs SFT.}
    \label{fig:13}
\end{figure}
\begin{figure}[h]
    \centering
    \includegraphics[width=0.8\linewidth]{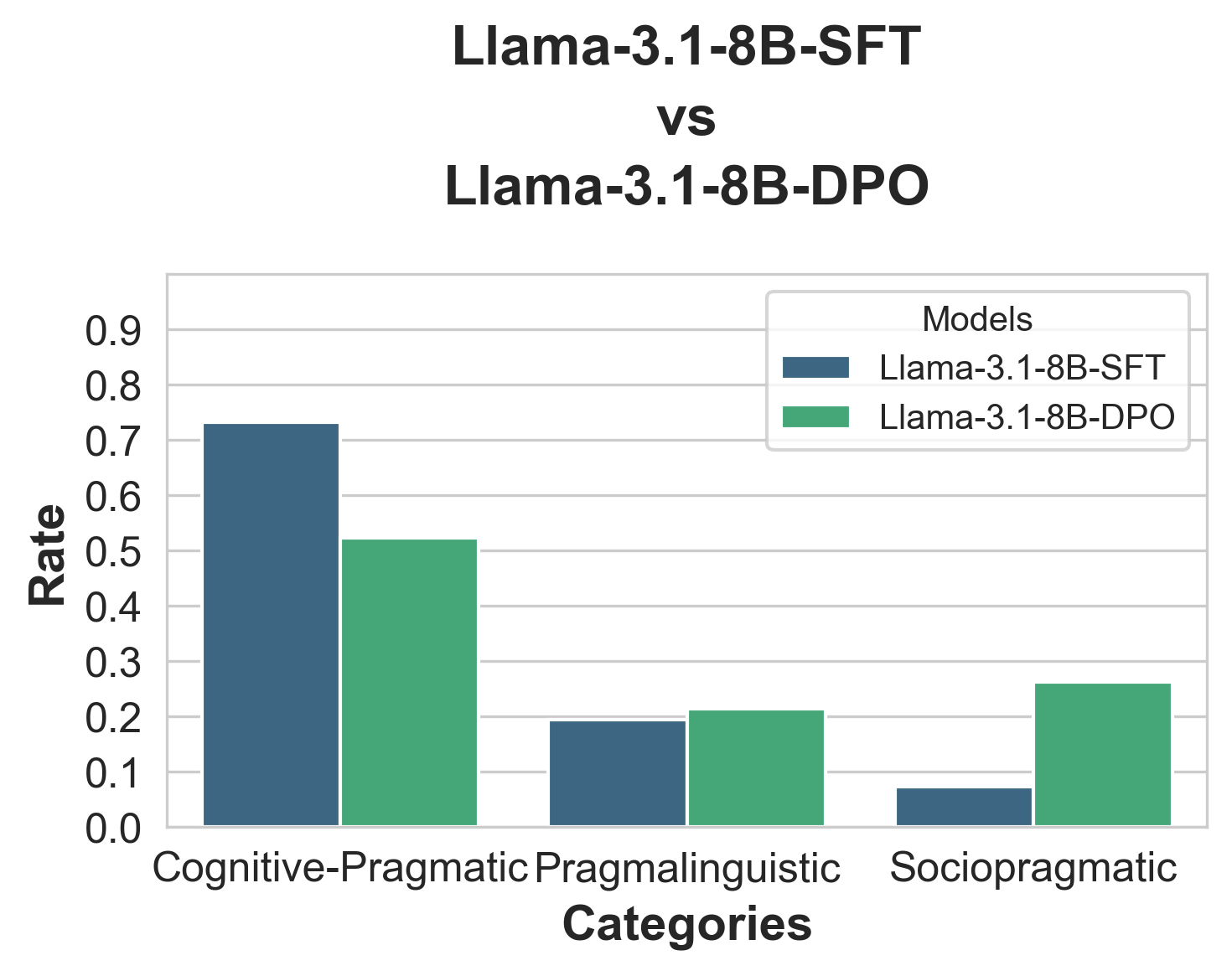}
    \caption{LLaMA-3.1-8B SFT vs DPO.}
    \label{fig:14}
\end{figure}
\begin{figure}[h]
    \centering
    \includegraphics[width=0.8\linewidth]{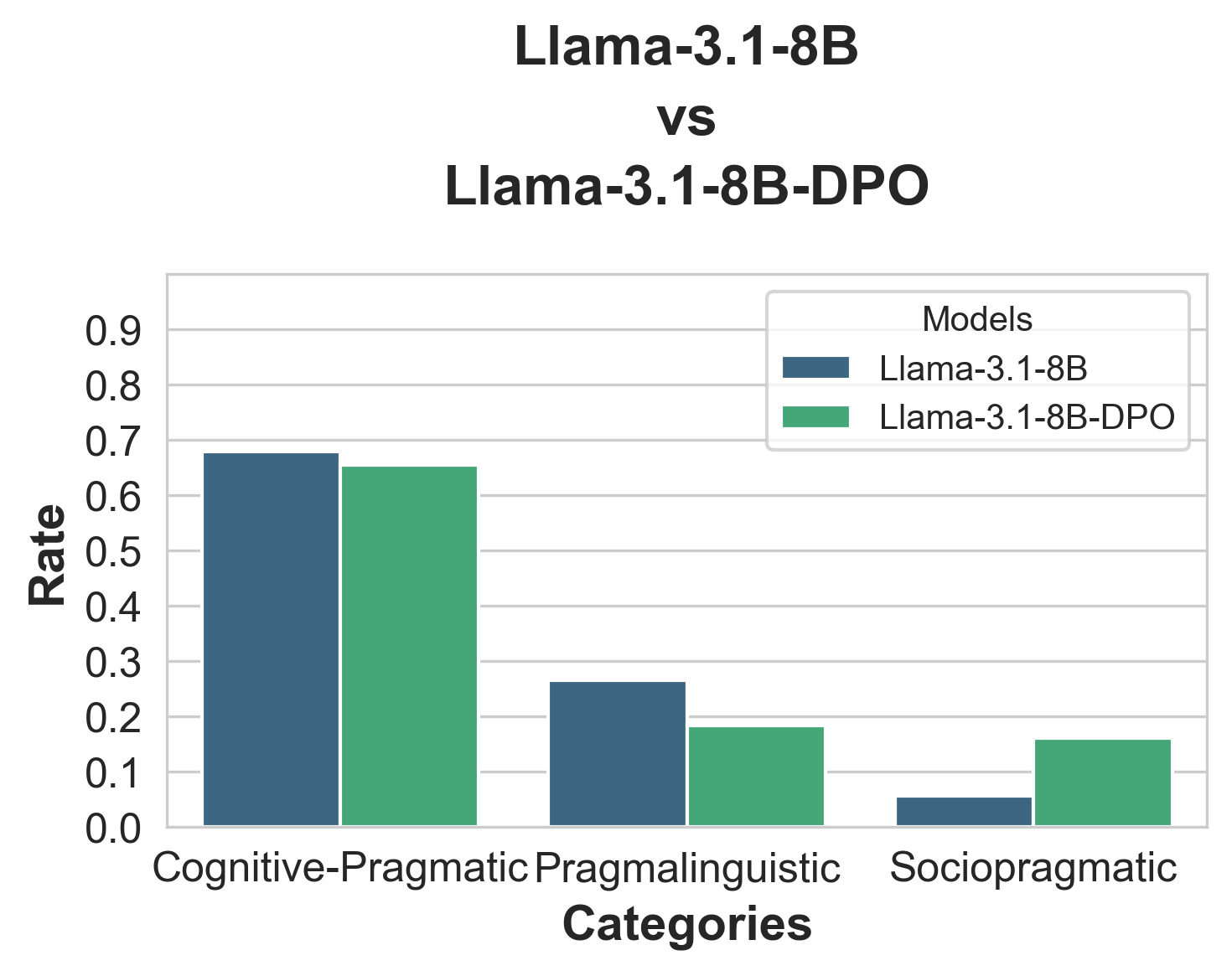}
    \caption{LLaMA-3.1-8B Base vs DPO.}
    \label{fig:15}
\end{figure}
\begin{figure}[h]
    \centering
    \includegraphics[width=0.8\linewidth]{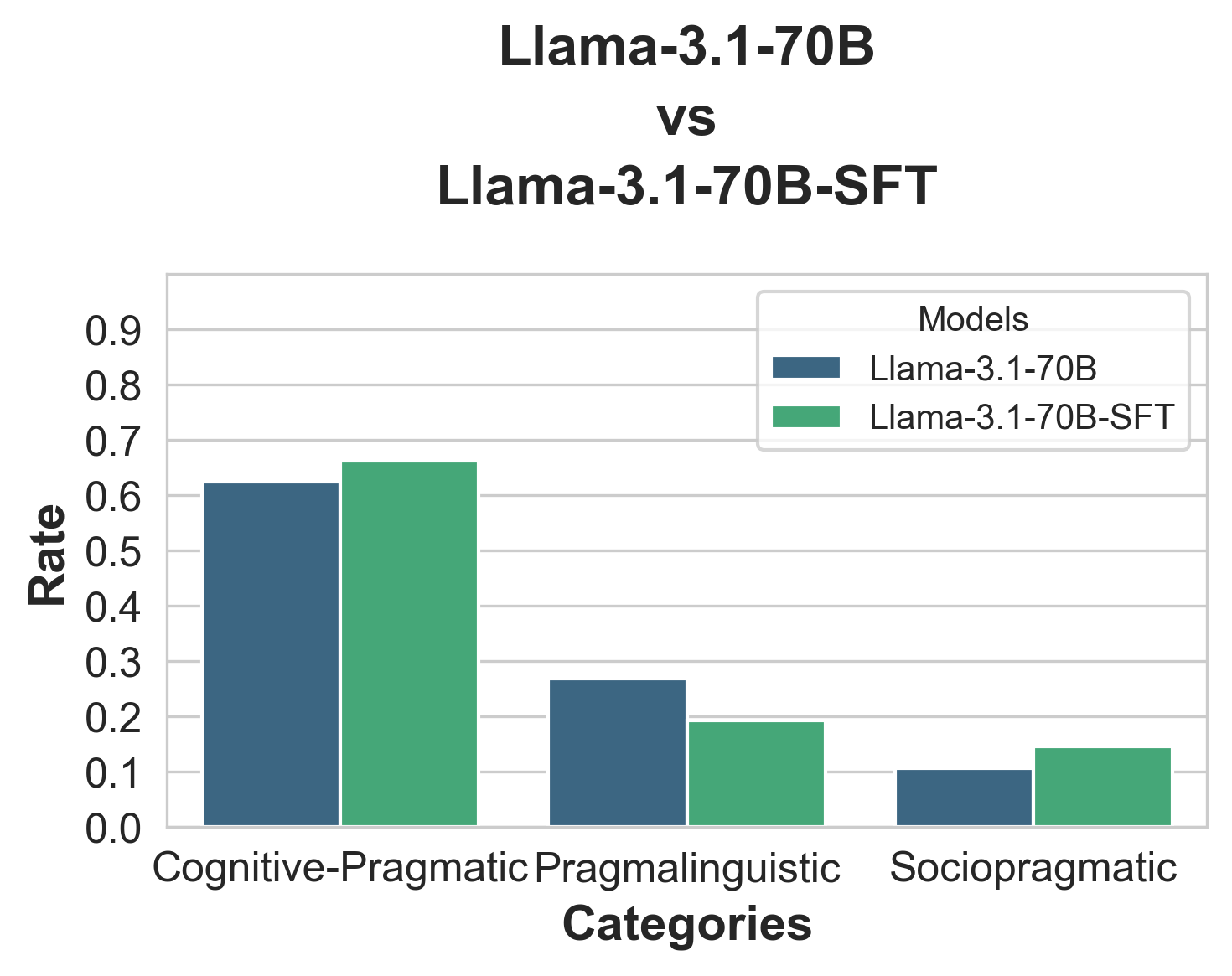}
    \caption{LLaMA-3.1-70B Base vs SFT.}
    \label{fig:16}
\end{figure}
\begin{figure}[h]
    \centering
    \includegraphics[width=0.8\linewidth]{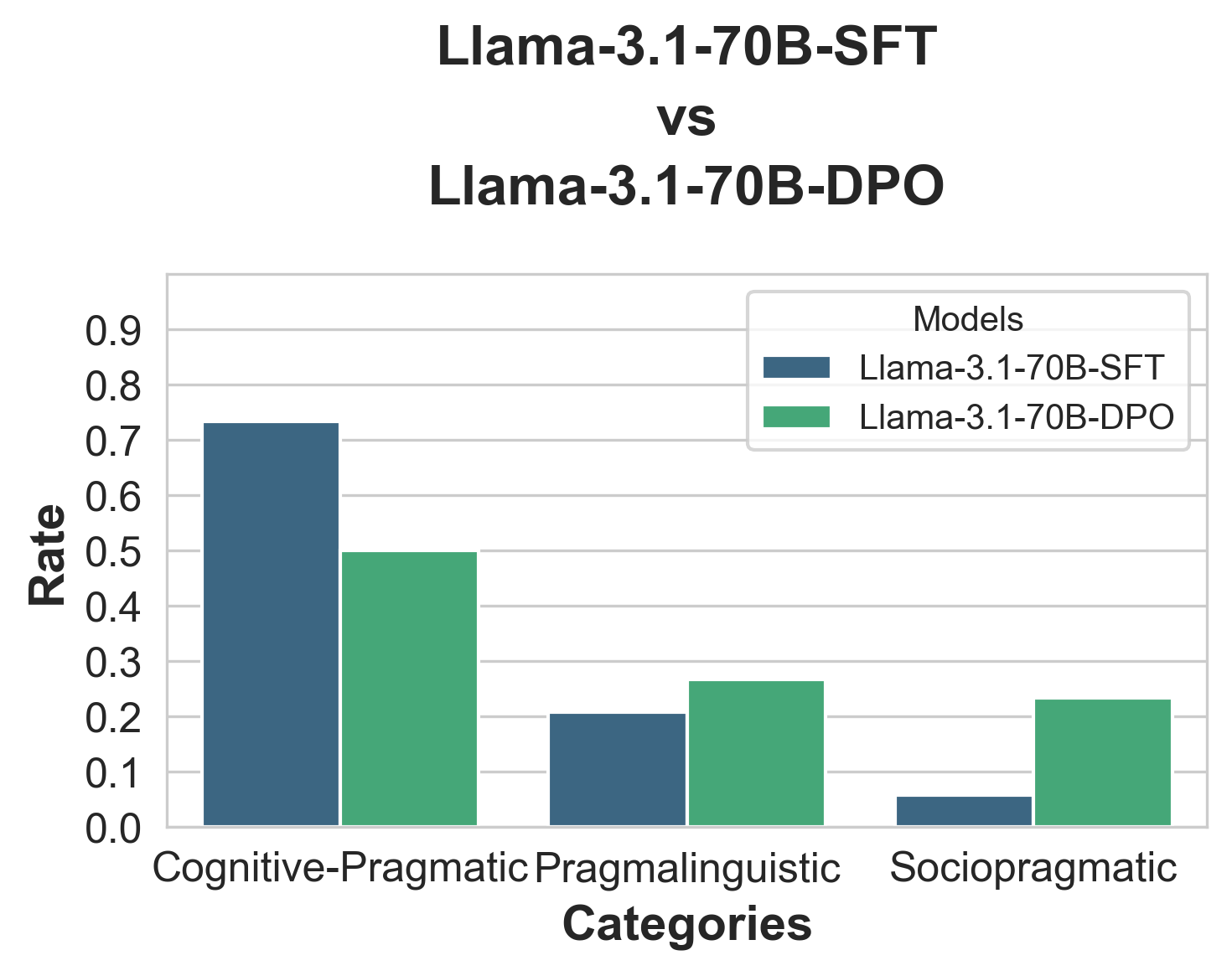}
    \caption{LLaMA-3.1-70B SFT vs DPO.}
    \label{fig:17}
\end{figure}
\begin{figure}[h]
    \centering
    \includegraphics[width=0.8\linewidth]{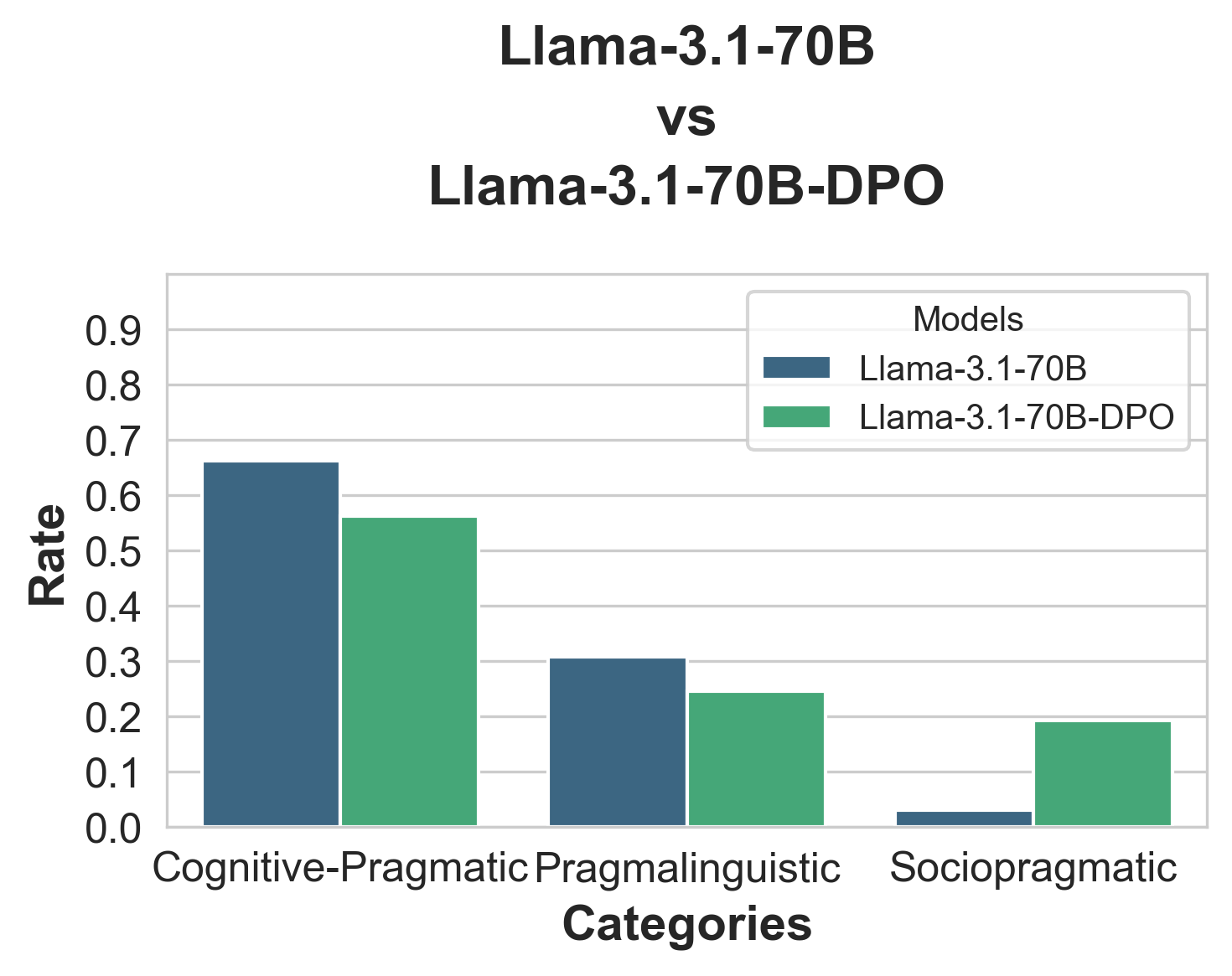}
    \caption{LLaMA-3.1-70B Base vs DPO.}
    \label{fig:18}
\end{figure}

\section{Generative AI Statement}
\label{appendix:ai-statement}

We use generative AI tools to assist with both the implementation and writing processes in this project. Specifically, we employed \textbf{Cursor}, an AI-assisted development environment, and \textbf{ChatGPT} (GPT-4o) to support the coding of evaluation tasks. Additionally, ChatGPT was used to aid in formatting sections of the paper, as well as generating LaTeX tables and figure templates. All outputs were carefully reviewed, edited, and verified by the authors to ensure factual accuracy and scholarly integrity.

\end{document}